\documentclass[11pt]{article}

\usepackage[final]{acl}

\usepackage{times}
\usepackage{latexsym}
\usepackage{amsmath}
\usepackage{amssymb}

\usepackage[T1]{fontenc}

\usepackage[utf8]{inputenc}

\usepackage{microtype}

\usepackage{graphicx}
\usepackage{placeins}

\usepackage{float}
\usepackage{booktabs}
\usepackage{multirow}
\usepackage{array}
\usepackage{adjustbox}
\usepackage{tikz}
\usepackage{dblfloatfix}
\usetikzlibrary{positioning,arrows.meta}
\usepackage[table]{xcolor}
\definecolor{simblue}{HTML}{2E5BB7}
\usepackage{threeparttable}
\usepackage{subcaption}
\usepackage{hyperref}
\hypersetup{
    colorlinks=true,
    linkcolor=blue,
    filecolor=magenta,      
    urlcolor=cyan,
}

\title{BuddyBench: A Privacy-Constrained Multi-Task Benchmark for Pediatric Social-Communication Personalization}

\author{
  \textbf{Jeyeon Eo}\textsuperscript{1},
  \textbf{Joo Young Kim}\textsuperscript{2},
  \textbf{Ran Ju}\textsuperscript{2},
  \textbf{Minyoung Jung}\textsuperscript{2},
  \textbf{Unggi Lee}\textsuperscript{3}\thanks{Corresponding author.}
\\[2pt]
  \textsuperscript{1}Independent Researcher, Seoul, Republic of Korea \\
  \textsuperscript{2}Neudive Inc., Daegu, Republic of Korea \\
  \textsuperscript{3}Korea University, Sejong, Republic of Korea \\[4pt]
  \small{\texttt{jeyeon.yona.eo@gmail.com},
    \texttt{jyjoshk@kitech.re.kr},
    \texttt{ran\_ju@neudive.com}}\\
  \small{\texttt{minyoung@kbri.re.kr},
    \texttt{codingchild@korea.ac.kr}}
}

\begin{document}
\maketitle

\begin{abstract}
    \textit{BuddyBench} introduces a privacy-constrained multi-task benchmark for pediatric social-communication personalization. Unlike existing neurodevelopmental repositories that primarily emphasize imaging, genetics, or cross-sectional clinical phenotyping, BuddyBench links drill-level learning trajectories, standardized clinical assessments, BuddyPlan self-report, and randomized-treatment endpoints within a unified benchmark schema. BuddyBench combines two cohorts: ND-03 is an observational cohort with dense drill coverage for Tasks~1--2 ($n = 189$), and ND-02 is a randomized controlled trial cohort for Tasks~3--4 ($n = 86$ ITT). Together, they support knowledge tracing, next-drill recommendation, clinical prediction, and causal inference, linking behavioral personalization to clinical evaluation. We additionally introduce \textit{BuddyBench-Sim}, a synthetic companion dataset for reproducible evaluation. Baselines show signal across tasks while keeping pediatric clinical records protected.
    \end{abstract}

\begin{figure*}
    \centering
    \vspace{-1em}
    \includegraphics[width=1\linewidth]{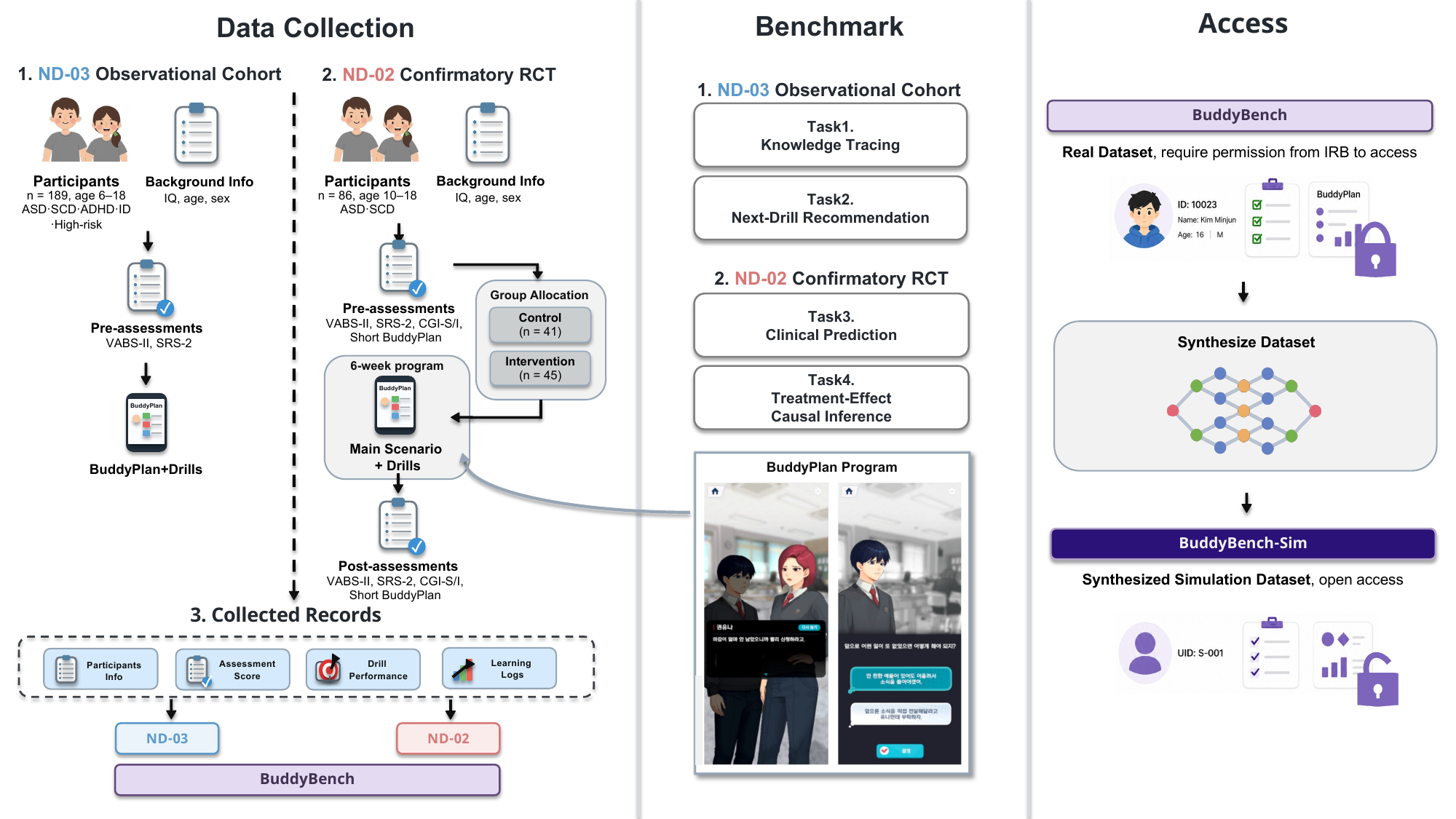}
    \caption{Overview of BuddyBench. The benchmark links behavioral Tasks~1--2 in ND-03 with clinically anchored Tasks~3--4 in ND-02, while BuddyBench-Sim enables open benchmarking under controlled real-data access.}
    \label{fig:buddybench-overview}
\end{figure*}

\begin{table*}[b]
    \centering
    \vspace{0.8em}
    \caption{Datasets closest to BuddyBench on benchmark-relevant dimensions.}
    \label{tab:dataset-comparison}
    \resizebox{\textwidth}{!}{%
    \begin{threeparttable}
    \vspace{-0.8em}
    \begin{tabular}{llcccccll}
    \toprule
    \textbf{Dataset} & \textbf{Population} & \textbf{Struct.\ Logs} & \textbf{Clin.\ Outcomes} & \textbf{Self-report} & \textbf{Skill Tax.} & \textbf{Benchmark} & \textbf{Size} & \textbf{Access} \\
    \midrule
    ABIDE I/II \citep{dimartino2014abide,dimartino2017abide2} & ASD & $\times$ & $\circ$ & $\circ$ & $\times$ & $\times$ & 2,156 & Public \\
    NDA \citep{nimhnda2026} & Neurodev & $\times$ & $\circ$ & $\circ$ & $\times$ & $\times$ & 25,000+ & Controlled \\
    SPARK \citep{spark2018cohort} & ASD + family & $\times$ & $\circ$ & $\checkmark$ & $\times$ & $\times$ & 50,000+ & Controlled \\
    HBN \citep{alexander2017hbn} & Pediatric MH & $\times$ & $\checkmark$ & $\checkmark$ & $\times$ & $\times$ & 10,000+ & Controlled \\
    MMASD \citep{li2023mmasd} & ASD intervention & $\circ$ & $\circ$ & $\times$ & $\circ$ & $\times$ & 32 & Public \\
    Engagnition \citep{kim2024engagnition} & ASD intervention & $\circ$ & $\times$ & $\times$ & $\times$ & $\times$ & 57 & Public \\
    \textbf{BuddyBench (ours)} & \textbf{ASD/SCD/high-risk} & $\checkmark$ & $\checkmark$ & $\checkmark$ & $\checkmark$ & $\checkmark$ & \textbf{269} & \textbf{Hybrid} \\
    \bottomrule
    \end{tabular}
    \begin{tablenotes}[flushleft]
    \footnotesize
    \item $\checkmark$: comprehensive; $\circ$: partial; $\times$: absent. Struct.\ Logs = item/drill-level records; Clin.\ Outcomes = validated endpoints beyond demographics/diagnosis; Skill Tax. = activity-to-skill mapping; Benchmark = released tasks, splits, metrics, and baselines. MH = Mental Health.
    \end{tablenotes}
    \end{threeparttable}%
    }
    \end{table*}

\section{Introduction}

Digital therapeutics for social-communication training in autism spectrum disorder (ASD) and related neurodevelopmental conditions are increasingly used as adjunct interventions, generating dense drill-level interaction logs alongside conventional clinical assessments \citep{daniels2018superpowerglass}. These logs are a natural substrate for personalization, supporting adaptive curricula, learner-state estimation, and treatment-response prediction. Yet no public benchmark connects intervention behavior to validated pediatric clinical outcomes within the same cohort.

Two adjacent research areas only partially address this gap. Educational AI benchmarks \citep{piech2015deep,choi2020ednet,liu2023xes3g5m} offer dense interaction logs but target math and science with no clinical endpoints; clinical prediction benchmarks \citep{johnson2023mimiciv,grundmann2026clinibench} carry validated outcomes but lack structured therapeutic logs. Existing neurodevelopmental datasets (Table~\ref{tab:dataset-comparison}) emphasize imaging or cross-sectional phenotyping, or provide behavioral signals without benchmark-grade splits and metrics \citep{dimartino2014abide,li2023mmasd}. On the causal side, few resources combine RCT-level randomization with structured behavioral logs \citep{wager2018estimation,nie2021rlearner}.

We introduce \textit{BuddyBench}\footnote{Access to the real cohorts requires IRB approval from both the requesting researcher's institution and the originating study's IRB. BuddyBench-Sim is openly available at the anonymous review repository: \url{https://anonymous.4open.science/r/BuddyBench-Sim}.}, a pediatric benchmark built from two 2024-2025 cohorts collected using the same social-communication training application. The first cohort, ND-03, is an \emph{observational} study of 189 children with dense drill-level interaction coverage and standardized clinical assessments. The second, ND-02, is a \emph{randomized controlled trial} of 86 children with treatment assignment and paired pre-/post-treatment clinical endpoints. Together the two cohorts share a 153-item drill bank, the BuddyPlan self-report instrument, and a communication skill taxonomy, enabling a unified benchmark interface.

BuddyBench defines four benchmark tasks across these two evidence layers as follows:

\textbf{Task 1: Knowledge tracing} (ND-03). Predict held-out per-drill correctness from a participant's prior drill history; primary metric AUC \citep{piech2015deep,xiong2016going}.

\textbf{Task 2: Next-drill recommendation} (ND-03). Rank plausible next drills given the observed practice prefix; primary metrics R@5, Recall@10, and NDCG@10 \citep{kang2018sasrec,sun2019bert4rec,he2020lightgcn}.

\textbf{Task 3: Clinical prediction} (ND-02). From leakage-safe pre-treatment covariates, estimate the risk that a participant will not improve across endpoints; primary metric AUPRC.

\textbf{Task 4: Causal inference} (ND-02). Under residualized R-loss, evaluate causal estimators of endpoint-level treatment effects in a small RCT \citep{wager2018estimation,kunzel2019metalearners,nie2021rlearner,shalit2017estimating}.

To support reproducible evaluation without redistributing pediatric clinical records, the real cohorts remain under IRB-gated controlled access. We release \textit{BuddyBench-Sim}, a schema-aligned synthetic counterpart whose utility and disclosure risk are jointly audited \citep{stadler2022synthetic,shokri2017membership}.

Our contributions are threefold:

\textbf{Benchmark}: 269 participants (183 ND-03 benchmark analysis set + 86 ND-02 ITT) across an observational and an RCT cohort, jointly aligning drill logs, BuddyPlan self-report, clinical assessments, and a skill taxonomy.

\textbf{Four tasks}: knowledge tracing, next-drill recommendation, clinical prediction, and causal inference, with conventional baselines across all tasks and LLM-based models for Task~1.

\textbf{Public artifact}: \textit{BuddyBench-Sim}, a schema-aligned synthetic counterpart that opens benchmarking while real cohorts remain under IRB-gated controlled access.

\begin{table*}[!t]
  \centering
  \small
  \setlength{\tabcolsep}{4pt}
  \renewcommand{\arraystretch}{0.92}
  \caption{Cohort characteristics and data coverage in BuddyBench.}
  \label{tab:cohort-overview}
  \vspace{-0.5em}
  \begin{tabular}{
  @{}p{0.19\textwidth}
  p{0.38\textwidth}
  p{0.34\textwidth}@{}
  }
  \toprule
  \textbf{Characteristic} & \textbf{ND-03 observational cohort} & \textbf{ND-02 RCT cohort} \\
  \midrule
  Participants & 189 subjects & 86 randomized subjects: 41 treatment / 45 control \\
  Collection setting & Multi-institution observational deployment & Multi-institution randomized controlled trial \\
  Analysis sets & Single released analysis set ($n{=}183$; 189 enrolled, 6 excluded for insufficient drill history) & ITT cohort ($n{=}86$); task-specific FAS/PP subsets where applicable \\
  Age range & 6-18 years & 10-18 years \\
  Diagnosis span & ASD, SCD, and additional clinical/comparison groups & ASD, SCD \\
  Drill bank & 153 items & 153 items \\
  BuddyPlan & 52-item instrument & 18-item subset \\
  Clinical measures & Baseline SRS-2/VABS-II and profile variables & Pre-/post-treatment SRS-2, VABS-II, BuddyPlan, and trial endpoint variables \\
  Drill-response coverage & 88.1\% observed drill-response matrix & 32.0\% observed drill-response matrix \\
  \bottomrule
  \end{tabular}
  \vspace{-0.5em}
\end{table*}

\begin{figure*}[!t]
\centering
\includegraphics[width=\textwidth]{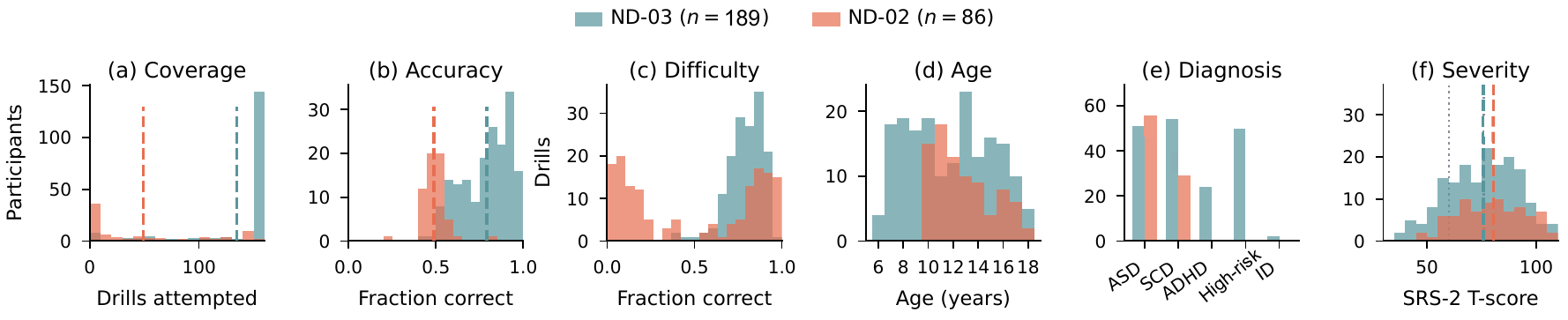}
\vspace{-2em}
\caption{Per-participant drill coverage and accuracy in BuddyBench. Dashed lines mark cohort means.}
\label{fig:data-distributions}
\end{figure*}

\section{Related Work}

\subsection{Clinical Datasets for Neurodevelopmental Intervention Research}

Large cohort resources such as ABIDE I/II \citep{dimartino2014abide,dimartino2017abide2} and HBN \citep{alexander2017hbn} support phenotyping and population-level analysis but lack drill-level interaction logs for benchmarking personalization methods. The closest intervention and behavioral comparators, MMASD \citep{li2023mmasd} and Engagnition \citep{kim2024engagnition}, provide richer multimodal behavioral signals in small ASD cohorts. MMASD covers play-therapy intervention recordings with pre-/post-test collection, while Engagnition focuses on serious-game engagement. Table~\ref{tab:dataset-comparison} positions BuddyBench against these resources.

\subsection{Task-Aligned Personalization Benchmarks}

BuddyBench combines four task families that are usually benchmarked separately. In educational AI, datasets such as XES3G5M \citep{liu2023xes3g5m} support knowledge tracing with rich auxiliary information, while Tenrec \citep{yuan2022tenrec} provides a large-scale multipurpose recommendation benchmark. Clinical prediction benchmarks such as CliniBench \citep{grundmann2026clinibench}, built on MIMIC-IV admission notes \citep{johnson2023mimiciv}, focus on hospital-text outcome prediction, while treatment-effect benchmarking in healthcare remains fragmented \citep{ling2023hte}. 
\subsection{Privacy-Preserving Clinical Data Release}

Synthetic data release has emerged as a third option alongside de-identification and controlled access. Prior work establishes that synthetic data must be assessed jointly for utility and disclosure risk \citep{stadler2022synthetic,woo2009pmse}. SDV, CTGAN, and TabDDPM provide generation methods for structured tabular data \citep{patki2016sdv,xu2019ctgan,kotelnikov2023tabddpm}, and membership inference attacks provide a concrete privacy audit for potential record memorization \citep{shokri2017membership}. BuddyBench adopts this approach, pairing controlled access for the real cohorts with the public BuddyBench-Sim release and reporting utility and privacy audits jointly.

    \begin{table*}[t]
        \centering
        \small
        \setlength{\tabcolsep}{4pt}
        \renewcommand{\arraystretch}{0.95}
        \caption{Benchmark task interface in BuddyBench}
        \label{tab:task-overview}
        \resizebox{\textwidth}{!}{%
        \begin{tabular}{
        @{}p{0.22\textwidth}
        p{0.09\textwidth}
        p{0.08\textwidth}
        p{0.25\textwidth}
        p{0.21\textwidth}
        p{0.09\textwidth}@{}
        }
        \toprule
        \textbf{Task} & \textbf{Cohort} & \textbf{$n$} & \textbf{Input} & \textbf{Output}\\
        \midrule
        Task 1: Knowledge tracing & ND-03 & 183 & Masked F3 token sequence with auxiliary context, learner history, and target drill & Held-out target-drill correctness\\
        Task 2: Next-drill recommendation & ND-03 & 183 & Support history and learner profile & Ranked future drill candidates \\
        Task 3: Clinical prediction & ND-02 & 83 FAS & Pre-treatment covariates & Clinical non-response risk\\
        Task 4: Causal inference & ND-02 & 86 ITT & Baseline covariates and treatment assignment & Endpoint-level treatment-effect estimate\\
        \bottomrule
        \end{tabular}}
        \vspace{-0.5em}
        \end{table*}
    
    
    \section{BuddyBench Dataset}
    
    \subsection{Cohort Design and Clinical Setting}
    
    BuddyBench integrates two pediatric cohorts collected using the same social-communication training application. ND-03 is the observational cohort for Tasks~1--2 with dense drill coverage ($n = 189$ enrolled; $n = 183$ in the benchmark analysis set after excluding 6 participants with insufficient drill history; October 2024--July 2025). ND-02 is the randomized controlled trial cohort for Tasks~3--4 ($n = 86$ ITT / 83 FAS / 75 PP). Both cohorts contain drill-level behavioral logs, BuddyPlan self-report items, demographic variables, and standardized clinical assessments. Table~\ref{tab:cohort-overview} summarizes cohort design and coverage. Figure~\ref{fig:buddybench-overview} shows how both cohorts map to the benchmark interface.
    
    \subsection{Representation and Measurement}
    
    BuddyBench aligns three modalities around a pseudonymous participant identifier. The participant table records cohort membership, demographic and diagnostic fields, baseline standardized measures, and study-specific profile variables. The behavioral matrix records binary per-drill correctness for the shared 153-item drill bank (\texttt{D001}-\texttt{D153}), with within-participant attempt order serving as the temporal signal for sequential models. The measurement tables cover SRS-2 \citep{constantino2012srs2} and VABS-II \citep{sparrow2016vabs3} scores in both cohorts, the 52-item BuddyPlan in ND-03 (18-item subset in ND-02), and CGI-S/I \citep{guy1976ecdeu} plus trial endpoints in ND-02.
    
    Each drill belongs to a shared social-communication training schema, allowing models to use item identity, learner history, and skill-level summaries without requiring raw natural-language session content. ND-02 additionally contains treatment assignment and paired pre-/post-treatment endpoint variables; these fields are retained for treatment-effect evaluation but excluded from predictive feature tiers whenever they would reveal post-treatment outcomes.
    
    \subsection{Cohort and Interaction Statistics}
    
    ND-03 spans a broader clinical and developmental range: SCD ($n = 54$), ASD ($n = 51$), high-risk group ($n = 50$), ADHD ($n = 24$), ID ($n = 2$), and two other or missing diagnostic codes. ND-02 is concentrated on ASD/SCD cases (ASD $n = 56$, SCD $n = 29$, one missing code) and has a balanced treatment allocation for a small RCT (41 treatment, 45 control). Sex ratios are 2.4:1 M:F in ND-03 and 13:1 M:F in ND-02, characteristic of ASD/SCD clinical trial populations. This composition gives BuddyBench broad behavioral variation for personalization tasks and randomized clinical structure for outcome-oriented evaluation.
    
    The interaction statistics show why the two cohorts support different benchmark roles. ND-03 has an 88.1\% observed drill-response fill rate, with participants attempting 137.4 drills on average and item-level accuracy ranging from 0.380 to 0.963. This gives the behavioral tasks genuine variation in both learner history and item difficulty. ND-02 has a 32.0\% fill rate and 49.9 attempts per participant on average, reflecting the narrower RCT protocol; its item-level accuracy spans a wider range, as small per-item sample sizes produce more extreme estimates. Figure~\ref{fig:data-distributions} visualizes these differences in participant-level coverage and accuracy.
    
    \subsection{Collection and Preprocessing}
    
    Drill responses are logged automatically during app-based practice sessions, while questionnaire and clinical assessments are administered under the corresponding observational or RCT protocol. Calendar timestamps and direct identifiers are removed; pseudonymous participant IDs and within-participant attempt indices are retained. Unattempted participant-drill cells are treated as missing observations rather than incorrect responses. For ND-02 prediction tasks, post-treatment outcomes, change scores, and variables derived from endpoint definitions are excluded from pre-treatment feature tiers.

    \providecommand{\modelcell}[1]{\begin{tabular}[t]{@{}l@{}}\texttt{#1}\\[-0.15em]\phantom{0}\end{tabular}}
    \providecommand{\modelcellbf}[1]{\begin{tabular}[t]{@{}l@{}}\textbf{\texttt{#1}}\\[-0.15em]\phantom{0}\end{tabular}}
    \providecommand{\metriccell}[2]{\begin{tabular}[t]{@{}c@{}}#1\\[-0.15em]\textcolor{simblue}{#2}\end{tabular}}
    \begin{table*}[b]
    \centering
    \scriptsize
    \setlength{\tabcolsep}{1.6pt}
    \renewcommand{\arraystretch}{0.86}
    \caption{Task~1 knowledge tracing under the recommended \texttt{F3} regime (5-fold CV, subject-level split); each metric cell shows BuddyBench (black) on the first line and BuddyBench-Sim (\textcolor{simblue}{blue}) on the second.}
    \label{tab:t1-kt-main}
    \vspace{-0.6em}

    \begin{adjustbox}{max width=\textwidth}
    \begin{tabular}{@{}lccc!{\vrule}lccc!{\vrule}lccc@{}}
    \toprule
    \textbf{Model} & \textbf{ACC}$\uparrow$ & \textbf{AUC}$\uparrow$ & \textbf{F1}$\uparrow$ &
    \textbf{Model} & \textbf{ACC}$\uparrow$ & \textbf{AUC}$\uparrow$ & \textbf{F1}$\uparrow$ &
    \textbf{Model} & \textbf{ACC}$\uparrow$ & \textbf{AUC}$\uparrow$ & \textbf{F1}$\uparrow$ \\
    \midrule

    \modelcell{AKT} & \metriccell{$0.79 \pm 0.03$}{$0.54 \pm 0.04$} & \metriccell{$0.65 \pm 0.02$}{$0.55 \pm 0.04$} & \metriccell{$0.88 \pm 0.02$}{$0.60 \pm 0.05$} &
    \modelcell{Hawkes} & \metriccell{$0.77 \pm 0.03$}{$0.51 \pm 0.01$} & \metriccell{$0.56 \pm 0.03$}{$0.51 \pm 0.01$} & \metriccell{$0.87 \pm 0.02$}{$0.48 \pm 0.04$} &
    \modelcell{DKVMN} & \metriccell{$0.79 \pm 0.03$}{$0.51 \pm 0.01$} & \metriccell{$0.65 \pm 0.04$}{$0.51 \pm 0.01$} & \metriccell{$0.88 \pm 0.02$}{$0.48 \pm 0.06$} \\

    \modelcell{HCGKT} & \metriccell{$0.79 \pm 0.03$}{$\mathbf{0.58 \pm 0.01}$} & \metriccell{$0.66 \pm 0.06$}{$0.61 \pm 0.01$} & \metriccell{$0.88 \pm 0.02$}{$0.57 \pm 0.01$} &
    \modelcell{IEKT} & \metriccell{$0.79 \pm 0.03$}{$0.58 \pm 0.01$} & \metriccell{$\underline{0.71 \pm 0.02}$}{$0.62 \pm 0.00$} & \metriccell{$0.88 \pm 0.02$}{$0.57 \pm 0.02$} &
    \modelcell{DKT+} & \metriccell{$0.79 \pm 0.03$}{$0.57 \pm 0.00$} & \metriccell{$0.65 \pm 0.03$}{$0.59 \pm 0.01$} & \metriccell{$0.88 \pm 0.02$}{$0.57 \pm 0.01$} \\

    \modelcell{KnowNet} & \metriccell{$0.79 \pm 0.03$}{$0.58 \pm 0.01$} & \metriccell{$0.66 \pm 0.02$}{$\underline{0.64 \pm 0.01}$} & \metriccell{$0.88 \pm 0.02$}{$\mathbf{0.58 \pm 0.03}$} &
    \modelcell{DKT} & \metriccell{$0.79 \pm 0.03$}{$0.57 \pm 0.00$} & \metriccell{$0.64 \pm 0.04$}{$0.59 \pm 0.01$} & \metriccell{$0.88 \pm 0.02$}{$0.57 \pm 0.01$} &
    \modelcell{KQN} & \metriccell{$0.79 \pm 0.03$}{$0.57 \pm 0.01$} & \metriccell{$0.62 \pm 0.03$}{$0.57 \pm 0.00$} & \metriccell{$0.88 \pm 0.02$}{$0.57 \pm 0.02$} \\

    \modelcell{GKT} & \metriccell{$0.79 \pm 0.03$}{$0.58 \pm 0.00$} & \metriccell{$0.62 \pm 0.03$}{$0.58 \pm 0.01$} & \metriccell{$0.88 \pm 0.02$}{$0.57 \pm 0.01$} &
    \modelcell{LEFOKT-AKT} & \metriccell{$0.79 \pm 0.03$}{$0.50 \pm 0.01$} & \metriccell{$0.63 \pm 0.02$}{$0.51 \pm 0.02$} & \metriccell{$0.88 \pm 0.02$}{$0.53 \pm 0.11$} &
    \modelcellbf{PEBG} & \metriccell{$0.79 \pm 0.03$}{$0.58 \pm 0.01$} & \metriccell{$\mathbf{0.72 \pm 0.03}$}{$0.63 \pm 0.01$} & \metriccell{$0.88 \pm 0.02$}{$0.57 \pm 0.02$} \\

    \modelcell{LOKT} & \metriccell{$0.86 \pm 0.02$}{$0.49 \pm 0.06$} & \metriccell{$0.70 \pm 0.05$}{$0.54 \pm 0.07$} & \metriccell{$0.92 \pm 0.02$}{$0.30 \pm 0.09$} &
    \modelcell{QDKT} & \metriccell{$0.79 \pm 0.02$}{$\underline{0.58 \pm 0.01}$} & \metriccell{$0.68 \pm 0.01$}{$\mathbf{0.64 \pm 0.01}$} & \metriccell{$0.88 \pm 0.02$}{$0.57 \pm 0.01$} &
    \modelcell{Thinking-KT} & \metriccell{$0.81 \pm 0.04$}{$0.46 \pm 0.07$} & \metriccell{$0.70 \pm 0.05$}{$0.54 \pm 0.07$} & \metriccell{$0.89 \pm 0.03$}{$0.13 \pm 0.06$} \\

    \modelcell{HISE-KT} & \metriccell{$\mathbf{0.86 \pm 0.03}$}{$0.49 \pm 0.07$} & \metriccell{$0.71 \pm 0.05$}{$0.55 \pm 0.07$} & \metriccell{$\mathbf{0.92 \pm 0.02}$}{$0.36 \pm 0.09$} &
    \modelcell{ReKT} & \metriccell{$0.79 \pm 0.03$}{$0.51 \pm 0.01$} & \metriccell{$0.60 \pm 0.04$}{$0.52 \pm 0.02$} & \metriccell{$0.88 \pm 0.02$}{$0.31 \pm 0.28$} &
    \modelcell{L-HAKT} & \metriccell{$\underline{0.86 \pm 0.03}$}{$0.49 \pm 0.07$} & \metriccell{$0.70 \pm 0.05$}{$0.54 \pm 0.07$} & \metriccell{$\underline{0.92 \pm 0.02}$}{$0.27 \pm 0.07$} \\

    \modelcell{RKT} & \metriccell{$0.79 \pm 0.03$}{$0.50 \pm 0.01$} & \metriccell{$0.68 \pm 0.03$}{$0.51 \pm 0.01$} & \metriccell{$0.88 \pm 0.02$}{$0.31 \pm 0.33$} &
    \modelcell{CIKT} & \metriccell{$0.85 \pm 0.02$}{$0.47 \pm 0.07$} & \metriccell{$0.71 \pm 0.05$}{$0.54 \pm 0.06$} & \metriccell{$0.92 \pm 0.02$}{$0.21 \pm 0.07$} &
    \modelcell{RobustKT} & \metriccell{$0.79 \pm 0.03$}{$0.51 \pm 0.04$} & \metriccell{$0.56 \pm 0.05$}{$0.52 \pm 0.04$} & \metriccell{$0.88 \pm 0.02$}{$0.26 \pm 0.22$} \\

    \modelcell{AKT-Custom} & \metriccell{$0.79 \pm 0.03$}{$0.56 \pm 0.03$} & \metriccell{$0.65 \pm 0.03$}{$0.56 \pm 0.04$} & \metriccell{$0.88 \pm 0.02$}{$0.48 \pm 0.18$} &
    \modelcell{SAINT} & \metriccell{$0.79 \pm 0.03$}{$0.57 \pm 0.01$} & \metriccell{$0.66 \pm 0.03$}{$0.61 \pm 0.01$} & \metriccell{$0.88 \pm 0.02$}{$\underline{0.58 \pm 0.01}$} &
    \modelcell{ATDKT} & \metriccell{$0.79 \pm 0.03$}{$0.57 \pm 0.01$} & \metriccell{$0.67 \pm 0.03$}{$0.59 \pm 0.01$} & \metriccell{$0.88 \pm 0.02$}{$0.56 \pm 0.01$} \\

    \modelcell{SAINT+} & \metriccell{$0.79 \pm 0.03$}{$0.57 \pm 0.00$} & \metriccell{$0.66 \pm 0.04$}{$0.61 \pm 0.01$} & \metriccell{$0.88 \pm 0.02$}{$0.57 \pm 0.00$} &
    \modelcell{ATKT} & \metriccell{$0.79 \pm 0.03$}{$0.58 \pm 0.00$} & \metriccell{$0.65 \pm 0.04$}{$0.59 \pm 0.01$} & \metriccell{$0.88 \pm 0.02$}{$0.57 \pm 0.01$} &
    \modelcell{SAKT} & \metriccell{$0.79 \pm 0.03$}{$0.50 \pm 0.01$} & \metriccell{$0.57 \pm 0.03$}{$0.50 \pm 0.01$} & \metriccell{$0.88 \pm 0.02$}{$0.41 \pm 0.30$} \\

    \modelcell{ATKTFix} & \metriccell{$0.76 \pm 0.03$}{$0.52 \pm 0.01$} & \metriccell{$0.58 \pm 0.03$}{$0.53 \pm 0.01$} & \metriccell{$0.86 \pm 0.02$}{$0.55 \pm 0.03$} &
    \modelcell{SimpleKT} & \metriccell{$0.79 \pm 0.03$}{$0.50 \pm 0.01$} & \metriccell{$0.68 \pm 0.03$}{$0.51 \pm 0.02$} & \metriccell{$0.88 \pm 0.02$}{$0.35 \pm 0.29$} &
    \modelcell{CSKT} & \metriccell{$0.79 \pm 0.03$}{$0.56 \pm 0.03$} & \metriccell{$0.67 \pm 0.03$}{$0.56 \pm 0.03$} & \metriccell{$0.88 \pm 0.02$}{$0.46 \pm 0.26$} \\

    \modelcell{SKVMN} & \metriccell{$0.79 \pm 0.03$}{$0.50 \pm 0.01$} & \metriccell{$0.54 \pm 0.01$}{$0.51 \pm 0.00$} & \metriccell{$0.88 \pm 0.02$}{$0.43 \pm 0.26$} &
    \modelcell{DTransformer} & \metriccell{$0.78 \pm 0.03$}{$0.50 \pm 0.01$} & \metriccell{$0.69 \pm 0.04$}{$0.50 \pm 0.01$} & \metriccell{$0.88 \pm 0.02$}{$0.34 \pm 0.19$} &
    \modelcell{StableKT} & \metriccell{$0.79 \pm 0.03$}{$0.57 \pm 0.01$} & \metriccell{$0.68 \pm 0.03$}{$0.60 \pm 0.01$} & \metriccell{$0.88 \pm 0.02$}{$0.56 \pm 0.01$} \\

    \modelcell{ExTRAKT} & \metriccell{$0.79 \pm 0.03$}{$0.50 \pm 0.01$} & \metriccell{$0.69 \pm 0.03$}{$0.51 \pm 0.01$} & \metriccell{$0.88 \pm 0.02$}{$0.30 \pm 0.30$} &
    \modelcell{UKT} & \metriccell{$0.79 \pm 0.03$}{$0.58 \pm 0.00$} & \metriccell{$0.61 \pm 0.06$}{$0.58 \pm 0.01$} & \metriccell{$0.88 \pm 0.02$}{$0.57 \pm 0.01$} &
    \modelcell{Deep-IRT} & \metriccell{$0.79 \pm 0.03$}{$0.50 \pm 0.01$} & \metriccell{$0.62 \pm 0.03$}{$0.50 \pm 0.01$} & \metriccell{$0.88 \pm 0.02$}{$0.50 \pm 0.07$} \\

    \bottomrule
    \end{tabular}
    \end{adjustbox}
    \vspace{-0.8em}
    \end{table*}

    \subsection{Benchmark Tasks}
    \label{sec:benchmark-tasks}
    
    BuddyBench pairs the cohorts with four task interfaces. T1 and T2 use dense ND-03 histories for learner-state modeling and recommendation, while T3 and T4 use ND-02 for pre-treatment risk prediction and treatment-effect evaluation. The tasks share a common schema but are not directly comparable because the two cohorts serve different roles.
    
    \textbf{Task 1: Knowledge tracing.} Given a participant's drill history in ND-03 and a target drill interaction, the model predicts held-out correctness.
    
    \textbf{Task 2: Next-drill recommendation.} Given the observed practice prefix in ND-03, the model ranks plausible next drills. This is an offline evaluation of curriculum consistency over reconstructed within-participant order, not a direct estimate of therapeutic efficacy.
    
    \textbf{Task 3: Clinical prediction.} Given leakage-safe pre-treatment covariates from ND-02, the model estimates the risk of clinical non-response. A participant is labeled as a non-responder if none of the three endpoints shows improvement from baseline to endpoint.
    
    \textbf{Task 4: Causal inference.} Given ND-02 baseline covariates, treatment assignment, and endpoint outcomes, the model estimates endpoint-level treatment effects in a small-sample RCT stress test.
    
    Table~\ref{tab:task-overview} summarizes task inputs, outputs, cohorts, and primary metrics; the next section reports the experimental protocol and results.

    \subsection{BuddyBench-Sim Public Counterpart}
    
    BuddyBench-Sim is the public synthetic counterpart to BuddyBench. It includes ND-03 and ND-02 synthetic cohorts that mirror the benchmark interface without redistributing pediatric clinical records. It is generated with a physics-informed VAE; Section~\ref{sec:synthetic-readiness} summarizes its fidelity, utility, and privacy checks.
    
    \subsection{Access and Release Design}
    
\paragraph{Data access.}
The real cohorts (ND-03, ND-02) are not publicly released. Researchers who wish to access the real data must submit IRB approval from their own institution alongside a formal request to the originating study's IRB. Data transfer is contingent on approval from both IRBs and execution of a data use agreement with the originating institution. BuddyBench-Sim is available under CC~BY-NC~4.0 at the benchmark repository.

\providecommand{\modelcell}[1]{\begin{tabular}[t]{@{}l@{}}\texttt{#1}\\[-0.15em]\phantom{0}\end{tabular}}
\providecommand{\modelcellbf}[1]{\begin{tabular}[t]{@{}l@{}}\textbf{\texttt{#1}}\\[-0.15em]\phantom{0}\end{tabular}}
\providecommand{\metriccell}[2]{\begin{tabular}[t]{@{}c@{}}#1\\[-0.15em]\textcolor{simblue}{#2}\end{tabular}}

\begin{table*}[t!]
\centering
\scriptsize
\setlength{\tabcolsep}{1.6pt}
\renewcommand{\arraystretch}{0.86}
\caption{Task~2 next-drill recommendation under the warm-start \texttt{F3} protocol; each metric cell shows BuddyBench (black) on the first line and BuddyBench-Sim (\textcolor{simblue}{blue}) on the second.}
\label{tab:t2-recsys-main}
\vspace{-0.6em}

\begin{adjustbox}{max width=\textwidth}
\begin{tabular}{@{}lccc!{\vrule}lccc!{\vrule}lccc@{}}
\toprule
\textbf{Model} & \textbf{R@5}$\uparrow$ & \textbf{R@10}$\uparrow$ & \textbf{NDCG@10}$\uparrow$ &
\textbf{Model} & \textbf{R@5}$\uparrow$ & \textbf{R@10}$\uparrow$ & \textbf{NDCG@10}$\uparrow$ &
\textbf{Model} & \textbf{R@5}$\uparrow$ & \textbf{R@10}$\uparrow$ & \textbf{NDCG@10}$\uparrow$ \\
\midrule

\modelcell{FailureRate}
& \metriccell{$\underline{0.18 \pm 0.02}$}{$0.15 \pm 0.01$}
& \metriccell{$\underline{0.32 \pm 0.02}$}{$0.27 \pm 0.01$}
& \metriccell{$\underline{0.43 \pm 0.02}$}{$0.40 \pm 0.01$}
&
\modelcell{TS-Rec}
& \metriccell{$0.17 \pm 0.02$}{$0.14 \pm 0.01$}
& \metriccell{$0.31 \pm 0.02$}{$0.27 \pm 0.01$}
& \metriccell{$0.40 \pm 0.01$}{$0.37 \pm 0.01$}
&
\modelcell{CSRec}
& \metriccell{$\mathbf{0.19 \pm 0.03}$}{$\mathbf{0.21 \pm 0.01}$}
& \metriccell{$0.30 \pm 0.03$}{$\mathbf{0.36 \pm 0.02}$}
& \metriccell{$0.43 \pm 0.04$}{$\mathbf{0.51 \pm 0.02}$}
\\

\modelcell{Cold-Start}
& \metriccell{$0.13 \pm 0.03$}{$0.13 \pm 0.01$}
& \metriccell{$0.23 \pm 0.03$}{$0.24 \pm 0.01$}
& \metriccell{$0.30 \pm 0.01$}{$0.33 \pm 0.02$}
&
\modelcell{C3Rec}
& \metriccell{$0.16 \pm 0.05$}{$\underline{0.18 \pm 0.01}$}
& \metriccell{$0.26 \pm 0.05$}{$\underline{0.31 \pm 0.01}$}
& \metriccell{$0.36 \pm 0.05$}{$0.44 \pm 0.02$}
&
\modelcell{NILUS}
& \metriccell{$0.15 \pm 0.03$}{$0.18 \pm 0.01$}
& \metriccell{$0.26 \pm 0.03$}{$0.30 \pm 0.02$}
& \metriccell{$0.36 \pm 0.04$}{$\underline{0.47 \pm 0.03}$}
\\

\modelcell{SASRec}
& \metriccell{$0.16 \pm 0.02$}{$0.12 \pm 0.01$}
& \metriccell{$0.26 \pm 0.02$}{$0.27 \pm 0.01$}
& \metriccell{$0.35 \pm 0.03$}{$0.34 \pm 0.01$}
&
\modelcell{SIGMA}
& \metriccell{$0.15 \pm 0.04$}{$0.10 \pm 0.02$}
& \metriccell{$0.26 \pm 0.04$}{$0.24 \pm 0.02$}
& \metriccell{$0.36 \pm 0.03$}{$0.31 \pm 0.01$}
&
\modelcell{PreferDiff}
& \metriccell{$0.18 \pm 0.04$}{$0.15 \pm 0.01$}
& \metriccell{$0.31 \pm 0.04$}{$0.27 \pm 0.01$}
& \metriccell{$0.43 \pm 0.03$}{$0.39 \pm 0.01$}
\\

\modelcellbf{DiffDiv}
& \metriccell{$0.18 \pm 0.05$}{$0.14 \pm 0.01$}
& \metriccell{$\mathbf{0.32 \pm 0.05}$}{$0.27 \pm 0.01$}
& \metriccell{$\mathbf{0.43 \pm 0.04}$}{$0.39 \pm 0.01$}
&
\modelcell{BERT4Rec}
& \metriccell{$0.09 \pm 0.04$}{$0.15 \pm 0.01$}
& \metriccell{$0.17 \pm 0.04$}{$0.27 \pm 0.01$}
& \metriccell{$0.25 \pm 0.06$}{$0.40 \pm 0.01$}
&
\modelcell{LightGCN}
& \metriccell{$0.10 \pm 0.04$}{$0.15 \pm 0.01$}
& \metriccell{$0.18 \pm 0.04$}{$0.27 \pm 0.01$}
& \metriccell{$0.25 \pm 0.03$}{$0.40 \pm 0.01$}
\\

\modelcell{SimGCL}
& \metriccell{$0.06 \pm 0.02$}{$0.15 \pm 0.01$}
& \metriccell{$0.15 \pm 0.02$}{$0.27 \pm 0.01$}
& \metriccell{$0.22 \pm 0.04$}{$0.40 \pm 0.01$}
&
\modelcell{RecVAE}
& \metriccell{$0.11 \pm 0.02$}{$0.15 \pm 0.01$}
& \metriccell{$0.20 \pm 0.02$}{$0.27 \pm 0.01$}
& \metriccell{$0.29 \pm 0.02$}{$0.40 \pm 0.01$}
&
\multicolumn{4}{c}{}
\\

\modelcell{Disco}
& \metriccell{$0.16 \pm 0.03$}{$0.15 \pm 0.01$}
& \metriccell{$0.27 \pm 0.03$}{$0.27 \pm 0.01$}
& \metriccell{$0.40 \pm 0.03$}{$0.40 \pm 0.01$}
&
\modelcell{GRACE}
& \metriccell{$0.06 \pm 0.01$}{$0.05 \pm 0.01$}
& \metriccell{$0.10 \pm 0.01$}{$0.08 \pm 0.01$}
& \metriccell{$0.16 \pm 0.03$}{$0.13 \pm 0.01$}
&
\multicolumn{4}{c}{}
\\

\bottomrule
\end{tabular}
\end{adjustbox}
\vspace{-0.8em}
\end{table*}

\begin{table}[t]
\centering
\scriptsize
\setlength{\tabcolsep}{3pt}
\renewcommand{\arraystretch}{0.86}
\caption{Task~3 clinical prediction (BuddyBench: F2 pre-treatment-only features, $n{=}83$; BuddyBench-Sim: F3, $n{=}1{,}000$); each metric cell shows BuddyBench (black) on the first line and BuddyBench-Sim (\textcolor{simblue}{blue}) on the second.}
\label{tab:t3-clinical-main}
\vspace{-0.6em}

\begin{tabular}{@{}lccc@{}}
\toprule
\textbf{Model} & \textbf{AUPRC}$\uparrow$ & \textbf{AUC}$\uparrow$ & \textbf{Brier}$\downarrow$ \\
\midrule

\modelcell{elastic\_net}
& \metriccell{$0.81 \pm 0.05$}{$\underline{0.97 \pm 0.01}$}
& \metriccell{$0.56 \pm 0.03$}{$0.64 \pm 0.10$}
& \metriccell{$0.22 \pm 0.03$}{$\mathbf{0.04 \pm 0.00}$}
\\

\modelcell{lasso}
& \metriccell{$0.80 \pm 0.05$}{$0.97 \pm 0.01$}
& \metriccell{$0.54 \pm 0.03$}{$0.65 \pm 0.11$}
& \metriccell{$0.23 \pm 0.03$}{$0.04 \pm 0.00$}
\\

\modelcell{ridge}
& \metriccell{$0.77 \pm 0.05$}{$0.97 \pm 0.01$}
& \metriccell{$0.48 \pm 0.05$}{$0.64 \pm 0.11$}
& \metriccell{$0.26 \pm 0.04$}{$0.04 \pm 0.00$}
\\

\modelcell{logistic\_reg.}
& \metriccell{$0.76 \pm 0.06$}{$\mathbf{0.98 \pm 0.01}$}
& \metriccell{$0.52 \pm 0.12$}{$0.67 \pm 0.11$}
& \metriccell{$0.32 \pm 0.11$}{$\underline{0.05 \pm 0.01}$}
\\

\modelcell{tabboost}
& \metriccell{$\underline{0.84 \pm 0.06}$}{$0.98 \pm 0.00$}
& \metriccell{$\mathbf{0.61 \pm 0.11}$}{$0.70 \pm 0.06$}
& \metriccell{$\mathbf{0.19 \pm 0.03}$}{$0.04 \pm 0.00$}
\\

\modelcell{catboost}
& \metriccell{$0.82 \pm 0.10$}{$0.98 \pm 0.01$}
& \metriccell{$\underline{0.60 \pm 0.17}$}{$0.70 \pm 0.11$}
& \metriccell{$0.19 \pm 0.04$}{$0.04 \pm 0.00$}
\\

\modelcell{xgboost}
& \metriccell{$0.81 \pm 0.05$}{$0.98 \pm 0.01$}
& \metriccell{$0.55 \pm 0.12$}{$\underline{0.72 \pm 0.06}$}
& \metriccell{$0.21 \pm 0.04$}{$0.04 \pm 0.00$}
\\

\modelcell{random\_forest}
& \metriccell{$0.77 \pm 0.12$}{$0.97 \pm 0.01$}
& \metriccell{$0.50 \pm 0.21$}{$0.66 \pm 0.10$}
& \metriccell{$\underline{0.20 \pm 0.04}$}{$0.04 \pm 0.00$}
\\

\modelcell{lightgbm}
& \metriccell{$0.84 \pm 0.07$}{$0.98 \pm 0.01$}
& \metriccell{$0.61 \pm 0.17$}{$0.67 \pm 0.09$}
& \metriccell{$0.23 \pm 0.08$}{$0.04 \pm 0.00$}
\\

\modelcell{tabm}
& \metriccell{$\mathbf{0.86 \pm 0.05}$}{$0.98 \pm 0.01$}
& \metriccell{$0.60 \pm 0.11$}{$0.72 \pm 0.10$}
& \metriccell{$0.20 \pm 0.02$}{$0.05 \pm 0.01$}
\\

\modelcell{tabtransformer}
& \metriccell{$0.79 \pm 0.04$}{$0.98 \pm 0.01$}
& \metriccell{$0.50 \pm 0.05$}{$\mathbf{0.73 \pm 0.06}$}
& \metriccell{$0.20 \pm 0.02$}{$0.04 \pm 0.00$}
\\

\modelcell{saint}
& \metriccell{$0.78 \pm 0.08$}{$0.97 \pm 0.02$}
& \metriccell{$0.55 \pm 0.17$}{$0.61 \pm 0.14$}
& \metriccell{$0.29 \pm 0.08$}{$0.05 \pm 0.01$}
\\

\modelcell{tabdpt}
& \metriccell{$0.73 \pm 0.11$}{$0.98 \pm 0.01$}
& \metriccell{$0.41 \pm 0.18$}{$0.70 \pm 0.07$}
& \metriccell{$0.32 \pm 0.06$}{$0.04 \pm 0.00$}
\\

\modelcell{TabPFN}
& \metriccell{$0.74 \pm 0.09$}{$0.97 \pm 0.01$}
& \metriccell{$0.40 \pm 0.16$}{$0.61 \pm 0.07$}
& \metriccell{$0.20 \pm 0.02$}{$0.04 \pm 0.00$}
\\

\modelcell{ft\_transformer}
& \metriccell{$0.73 \pm 0.08$}{$0.97 \pm 0.02$}
& \metriccell{$0.38 \pm 0.14$}{$0.64 \pm 0.10$}
& \metriccell{$0.22 \pm 0.05$}{$0.05 \pm 0.01$}
\\

\bottomrule
\end{tabular}
\vspace{-0.8em}
\end{table}

\section{Experimental Setup and Results}

\subsection{Experimental Setup}

This section reports the main comparisons for Tasks~1--4. Tasks~1--4 use participant-split outer folds on the real-data cohorts.

The feature-ablation policy follows an F0--F3 scheme. F0 contains the minimal task inputs; F1 and F2 add task-legal demographic, baseline clinical, and cohort-specific questionnaire variables; F3 adds the richest leakage-safe augmentation, including drill-history summaries where applicable. The exact contents differ by task, but every tier is restricted to information available at the prediction time. Cross-task interpretation should therefore be made cautiously: T1 uses a tokenized masked-sequence F3 protocol, whereas T2--T4 use task-specific tabular feature tiers. The per-task model registry appears in Table~\ref{tab:model-registry}.

Leakage checks are especially important for ND-02: T3 excludes post-treatment endpoints, change scores, and variables derived from the outcome definition, whereas T4 uses treatment assignment and endpoint outcomes by design. Unless a caption says otherwise, a value reported as $a \pm b$ denotes mean $\pm$ fold-wise SD over the reported evaluations for that row; the uncertainty-reporting convention per task is summarized in Table~\ref{tab:uncertainty-ledger}. Hyperparameter tuning protocols are documented in Appendix~\ref{app:tuning}, and bootstrap / multiple-testing details in Appendix~\ref{app:statistics}.

\subsection{Results}
\subsubsection{Benchmark Behavior Across Tasks}
\textbf{Task 1: Knowledge tracing.}
Task~1 shows clear KT signal without collapsing into an easy benchmark (Table~\ref{tab:t1-kt-main}). On real ND-03, AUC spans 0.54--0.72 across the leaderboard; ACC and F1 converge near $0.79$ and $0.88$ for many neural models because predictions collapse to the majority class at the default threshold---AUC is therefore the primary discriminative metric (Appendix~\ref{app:t1-extended}, \ref{app:llm-kt-gap}). On BuddyBench-Sim, some ranking is preserved but several models remain near chance, showing that schema compatibility alone does not guarantee KT transfer.

\textbf{Task 2: Next-drill recommendation.}
Task~2 shows that next-drill recommendation is feasible but far from saturated under the warm-start protocol (Table~\ref{tab:t2-recsys-main}). Drill order is inferred from column position in the wide behavioral matrix, as calendar timestamps are withheld to reduce re-identification risk (Appendix~\ref{app:limitations}). Real-cohort ranking quality remains modest, and the gap between stronger learned recommenders and lightweight references is informative without suggesting that the task is solved. BuddyBench-Sim generally yields higher ranking scores, consistent with its larger and smoother interaction structure.

\begin{table}[t]
\centering
\scriptsize
\setlength{\tabcolsep}{3pt}
\renewcommand{\arraystretch}{0.86}
\caption{Task~4 causal inference under the residualized R-loss protocol; each metric cell shows BuddyBench (black) on the first line and BuddyBench-Sim (\textcolor{simblue}{blue}) on the second.}
\label{tab:t4-causal-main}
\vspace{-0.6em}

\begin{tabular}{@{}lccc@{}}
\toprule
\textbf{Model} & \textbf{CGI-S}$\downarrow$ & \textbf{SRS-2}$\downarrow$ & \textbf{VABS}$\downarrow$ \\
\midrule

\modelcell{ATEOnly}
& \metriccell{$\underline{0.25 \pm 0.06}$}{$\mathbf{2.44 \pm 0.59}$}
& \metriccell{$\mathbf{181.37 \pm 35.33}$}{$92.56 \pm 12.64$}
& \metriccell{$\underline{9.40 \pm 3.71}$}{$47.52 \pm 8.73$}
\\

\modelcell{ZeroITE}
& \metriccell{$0.25 \pm 0.06$}{$2.60 \pm 0.52$}
& \metriccell{$181.52 \pm 34.87$}{$94.36 \pm 13.98$}
& \metriccell{$10.00 \pm 4.05$}{$\underline{47.41 \pm 7.93}$}
\\

\modelcell{TTest}
& \metriccell{$0.25 \pm 0.06$}{$2.44 \pm 0.59$}
& \metriccell{$181.37 \pm 35.33$}{$92.56 \pm 12.64$}
& \metriccell{$9.40 \pm 3.71$}{$47.52 \pm 8.73$}
\\

\modelcell{ANCOVA}
& \metriccell{$0.25 \pm 0.06$}{$\underline{2.45 \pm 0.60}$}
& \metriccell{$\underline{181.43 \pm 35.71}$}{$93.06 \pm 12.95$}
& \metriccell{$9.48 \pm 3.80$}{$48.00 \pm 8.25$}
\\

\modelcell{OLSInteract.}
& \metriccell{$0.31 \pm 0.12$}{$3.92 \pm 0.57$}
& \metriccell{$212.47 \pm 39.01$}{$151.90 \pm 25.55$}
& \metriccell{$10.36 \pm 4.31$}{$88.17 \pm 12.30$}
\\

\modelcell{CausalForest}
& \metriccell{$\mathbf{0.24 \pm 0.06}$}{$2.44 \pm 0.57$}
& \metriccell{$182.26 \pm 36.87$}{$92.64 \pm 12.62$}
& \metriccell{$9.52 \pm 3.71$}{$\mathbf{47.33 \pm 8.18}$}
\\

\modelcell{DRLearner}
& \metriccell{$0.34 \pm 0.12$}{$3.87 \pm 0.71$}
& \metriccell{$211.61 \pm 76.78$}{$138.97 \pm 22.47$}
& \metriccell{$11.58 \pm 5.15$}{$85.00 \pm 14.77$}
\\

\modelcell{RLearner}
& \metriccell{$0.31 \pm 0.07$}{$3.73 \pm 0.53$}
& \metriccell{$183.21 \pm 56.58$}{$118.00 \pm 29.09$}
& \metriccell{$11.04 \pm 4.96$}{$78.19 \pm 12.42$}
\\

\modelcell{SLearner}
& \metriccell{$0.25 \pm 0.07$}{$2.48 \pm 0.58$}
& \metriccell{$184.03 \pm 36.24$}{$\underline{91.89 \pm 13.20}$}
& \metriccell{$10.08 \pm 3.96$}{$47.51 \pm 8.11$}
\\

\modelcell{TLearner}
& \metriccell{$0.34 \pm 0.12$}{$2.53 \pm 0.62$}
& \metriccell{$204.56 \pm 37.18$}{$92.27 \pm 10.82$}
& \metriccell{$11.92 \pm 4.26$}{$52.83 \pm 7.59$}
\\

\modelcell{XLearner}
& \metriccell{$0.31 \pm 0.10$}{$2.52 \pm 0.58$}
& \metriccell{$194.12 \pm 38.91$}{$\mathbf{90.56 \pm 11.26}$}
& \metriccell{$11.33 \pm 4.09$}{$48.59 \pm 6.22$}
\\

\modelcell{DragonNet}
& \metriccell{$0.34 \pm 0.13$}{$2.72 \pm 0.48$}
& \metriccell{$211.50 \pm 35.39$}{$104.87 \pm 18.13$}
& \metriccell{$11.54 \pm 4.99$}{$54.30 \pm 5.48$}
\\

\modelcell{TARNet}
& \metriccell{$0.34 \pm 0.14$}{$2.73 \pm 0.46$}
& \metriccell{$209.25 \pm 33.57$}{$107.37 \pm 20.10$}
& \metriccell{$10.92 \pm 4.93$}{$55.13 \pm 7.00$}
\\

\modelcell{BART}
& \metriccell{$0.26 \pm 0.07$}{$2.45 \pm 0.60$}
& \metriccell{$187.32 \pm 39.44$}{$92.54 \pm 12.05$}
& \metriccell{$\mathbf{9.39 \pm 3.65}$}{$47.43 \pm 8.66$}
\\

\modelcell{BCF}
& \metriccell{$0.25 \pm 0.06$}{$2.44 \pm 0.59$}
& \metriccell{$181.93 \pm 35.55$}{$92.80 \pm 12.72$}
& \metriccell{$9.43 \pm 3.74$}{$47.59 \pm 8.76$}
\\

\modelcell{CausalPFN}
& \metriccell{$0.32 \pm 0.12$}{$2.46 \pm 0.66$}
& \metriccell{$200.50 \pm 35.87$}{$92.68 \pm 14.34$}
& \metriccell{$12.14 \pm 5.46$}{$49.37 \pm 10.49$}
\\

\modelcell{GRFCausalForest}
& \metriccell{$0.25 \pm 0.05$}{$2.44 \pm 0.60$}
& \metriccell{$182.87 \pm 36.40$}{$92.17 \pm 12.54$}
& \metriccell{$9.59 \pm 3.75$}{$47.64 \pm 8.48$}
\\

\bottomrule
\end{tabular}
\vspace{-0.8em}
\end{table}

\textbf{Task 3: Clinical prediction.}
Task~3 shows that leakage-safe pre-treatment variables (F2 feature set) support non-trivial risk ranking, but not a deployment-ready screening pipeline (Table~\ref{tab:t3-clinical-main}). The non-response rate is 0.747 (62 of 83 FAS participants; Appendix~\ref{app:cohort}), establishing the prevalence-baseline AUPRC at 0.747; the best model (\texttt{tabm}, $0.86$) represents a $+0.11$ lift above this baseline. The stronger BuddyBench-Sim block should be read as executability evidence under a smoother synthetic regime, not as a match to real-RCT difficulty. A sample-size scaling analysis (Appendix~\ref{app:scaling}) confirms that the observed AUPRC reflects the data-limited regime rather than a model ceiling: subsampling the cohort to $n = 45$ reduces AUPRC by 15 points, and BuddyBench-Sim scale-up reaches $0.981$ at $n = 1000$, a performance level unreachable with real data alone.

\textbf{Task 4: Causal inference.}
Task~4 is best read as a methodological stress test under a small real RCT, not as a clinical-effect discovery claim (Table~\ref{tab:t4-causal-main}). Performance varies substantially by endpoint, and simple constant-effect references remain competitive on some outcomes; at $n{=}86$, no model is reliably differentiated from these references---the CGI-S gap between \texttt{CausalForest} and \texttt{ATEOnly} is $\Delta{=}0.009$ residualized R-loss (Appendix~\ref{app:difficulty}). The real and synthetic columns are therefore best interpreted as within-endpoint diagnostics. The residualized R-loss definition and the full causal-estimator leaderboard appear in Appendix~\ref{app:t4-extended} (with the R-loss derivation in Appendix~\ref{app:t4-rloss}). Appendix~\ref{app:scaling} further shows that SRS-2 R-loss decreases by 29\% as cohort size grows from $n=30$ to $n=75$, confirming the estimators are sample-starved rather than misspecified.

\subsubsection{Synthetic Data Fidelity, Utility, and Release Readiness}
\label{sec:synthetic-readiness}

BuddyBench-Sim is intended to support public benchmarking rather than to outperform the real cohorts. Because every field in the BuddyBench schema---diagnostic category, IQ, clinical scale scores, and drill response patterns---constitutes sensitive pediatric clinical information, re-identification risk rather than attribute inference is the central privacy concern; membership-inference accuracy and $k$-anonymity groupings therefore serve as the primary disclosure audit. The released PI-VAE synthetic cohorts preserve broad structure (pairwise feature-correlation alignment Spearman $\geq 0.968$; record-level clinical validity $=0.990$) and scale the public interface to 1{,}000 records. Their aggregate ML utility score is 0.780, membership-inference accuracy remains near chance ($\leq 0.52$), and the evaluated demographic--diagnostic--severity groupings satisfy $\min k \geq 5$. We therefore use BuddyBench-Sim as a public development interface rather than as a substitute for the real cohorts; the Task~3 synthetic-vs-real utility comparison appears in Table~\ref{tab:synthetic-t3}.
\section{Discussion}
\label{sec:discussion}

Across all four tasks, simple domain-aware baselines remain competitive with complex learned models, pointing to data scale rather than model capacity as the primary bottleneck. T1/T2 show learnable structure under the dense ND-03 logs; T3/T4 expose the harder floor set by the small ND-02 RCT and high class imbalance.

BuddyBench-Sim extends the benchmark to public experiments without redistributing pediatric clinical records, but its task-level behavior does not replicate real cohorts one-for-one. Some KT models remain near chance and graph-based recommenders gain substantially on Sim (e.g., \texttt{BERT4Rec} NDCG@10 $0.25 \to 0.40$) owing to its denser interaction structure, while frequency-based references remain flat; T3 AUPRC is inflated by synthetic prevalence and separability, and T4 reflects a different data-generating process with observable counterfactuals. Synthetic results should therefore be interpreted as method-development evidence.

For practitioners building toward clinical applications, BuddyBench tasks serve as method-screening evidence, not deployment readiness tests. A recommender achieving NDCG@10 of $0.43$ on T2 demonstrates curriculum-consistency modeling under an offline reconstructed protocol, not therapeutic benefit. A T3 classifier reaching AUPRC of $0.86$ ($+0.11$ above the prevalence baseline) identifies a statistically learnable risk-stratification signal; applying it in clinical triage would still require prospective validation, safety review, and regulatory assessment beyond offline benchmarking. BuddyBench is positioned upstream of those requirements, enabling reproducible method comparison at the behavioral-to-clinical interface while real-cohort access remains governed by IRB protocols and efficacy by prospective evaluation.

\section{Conclusion}

We introduce BuddyBench, a benchmark for social-communication personalization built from two pediatric cohorts with distinct roles: ND-03 for behavioral tasks and ND-02 for clinically anchored evaluation. Baselines confirm usable signal under leakage-aware protocols; the synthetic counterpart alters task difficulty rather than reproducing real-cohort behavior one-for-one. The main contribution is the integrated benchmark design and controlled/public release split, intended to support reproducible method comparison in privacy-constrained clinical settings.

\section{Ethical considerations}

BuddyBench is built from pediatric clinical data collected under institutional review board (IRB) oversight; data governance is aligned with the approved consent and ethical framework of each cohort.

\paragraph{Consent scope.}
ND-03 data were collected under an observational protocol approved for research analysis. ND-02 data were collected under an RCT protocol whose participant consent covers research analysis and regulatory/permit submissions; the consent does not extend to general AI model development or to unrestricted third-party redistribution. Accordingly, this paper presents ND-02 results as methodology benchmarking within the approved research scope, and does not position the real cohorts as a training corpus for external AI systems.

\paragraph{Real-data access governance.}
The real ND-02 and ND-03 cohorts are not publicly released. Researchers wishing to replicate or extend results on the real data must obtain IRB approval from both (i)~their own institution and (ii)~the originating study's IRB before any data transfer can be arranged. This dual-IRB requirement reflects the consent constraints described above and the elevated re-identification risk of the ND-02 cohort (small $n$, rare diagnostic population, extreme sex ratio; see Appendix~\ref{app:reidentification}).

\paragraph{BuddyBench-Sim and open benchmarking.}
BuddyBench-Sim, the synthetic counterpart, is released openly (CC~BY-NC~4.0) precisely to enable reproducible benchmarking without exposing consented clinical records. Researchers are encouraged to use BuddyBench-Sim for method development and preliminary evaluation prior to applying for real-data access.

\paragraph{Participant protection and clinical positioning.}
We position BuddyBench as a research resource for computational methodology rather than a clinical decision system. Per-subgroup performance breakdowns for ND-02 are suppressed to reduce re-identification risk, and no individual-level predictions or case-level outputs are released. Any translational application of methods developed on this benchmark would require independent prospective validation, safety review, and regulatory scrutiny beyond the scope of this work.

\paragraph{Assent and minor participants.}
All participants were minors; data collection followed protocols for pediatric research including parental consent and, where applicable, child assent procedures as required by the applicable IRB.

\section{Limitations}

BuddyBench is a single-site pediatric benchmark with modest cohort sizes, imbalanced sex ratios, and different coverage profiles across cohorts. ND-03 supports dense behavioral modeling, whereas ND-02 provides randomized structure but has sparse app-use coverage and small clinical analysis sets. The benchmark is therefore strongest as a method-comparison resource, not as population-representative evidence for subgroup behavior, clinical screening, or stable heterogeneous treatment effects. Appendix~\ref{app:limitations} elaborates these limitations, Appendix~\ref{app:datasheet} provides the dataset documentation, and Appendix~\ref{app:reproducibility} lists the reproducibility checklist; fold-level results, feature ablations (Appendix~\ref{app:feature-ablation}), and per-task extended leaderboards are collected in Appendix~\ref{app:extended-results}.

Fine-grained calendar time is omitted to reduce re-identification risk.

\section{GenAI Usage Disclosure}
GPT-5, 5.4, 5.5, and Claude Code were used for minor language editing and assistance with paper drafting. Qwen3-1.7B were used solely to run model predictions and evaluations. All scientific ideas, analyses, and experimental designs were conducted by the human authors.

\bibliography{custom}

\newpage

\clearpage

\appendix
\label{sec:appendix}
\clearpage

\section{Extended Baseline Results}
\label{app:extended-results}

Unless otherwise noted, feature-set labels follow the main-paper convention: F0-F2 are aligned shared ablation names, while F3 denotes the richest task-legal augmentation for each task rather than one globally identical column recipe. This section supplements the main paper with full leaderboards, feature-tier ablations, and BuddyBench-Sim diagnostics, without introducing new headline claims. Real and synthetic tables represent different difficulty levels and should not be treated as equivalent; the reported numbers clarify the benchmark behavior described in the main text.

\begin{table*}[!htbp]
\centering
\scriptsize
\setlength{\tabcolsep}{4pt}
\renewcommand{\arraystretch}{0.95}
\caption{Model registry for the reported benchmark families.}
\label{tab:model-registry}
\resizebox{\textwidth}{!}{%
\begin{tabular}{p{0.25\textwidth}p{0.08\textwidth}p{0.31\textwidth}p{0.28\textwidth}}
\toprule
\textbf{Model} & \textbf{Task} & \textbf{Family} & \textbf{Citation} \\
\midrule
DKT / DKT+ & T1 & Recurrent knowledge tracing & \citet{piech2015deep} \\
DKVMN / SKVMN & T1 & Memory-augmented knowledge tracing & \citet{zhang2017dkvmn} \\
SAKT / AKT / AKT-Custom & T1 & Attention-based knowledge tracing & \citet{pandey2019sakt,ghosh2020akt} \\
pyKT-registered KT variants & T1 & Standardized deep-KT benchmark implementations & \citet{liu2022pykt} \\
BPR / NMF / SVD / NCF & T2 & Matrix factorization and neural collaborative filtering & \citet{rendle2009bpr,he2017ncf} \\
SASRec / BERT4Rec & T2 & Sequential recommendation transformers & \citet{kang2018sasrec,sun2019bert4rec} \\
LightGCN / SimGCL & T2 & Graph-based and contrastive recommendation & \citet{he2020lightgcn,yu2022simgcl} \\
RecVAE & T2 & Variational-autoencoder recommendation & \citet{shenbin2020recvae} \\
CSRec / PreferDiff / C3Rec & T2 & Recent contrastive, diffusion, and course-recommendation baselines & \citet{liu2024csrec,liu2024preferdiff,li2025c3rec} \\
Elastic net / lasso / ridge / logistic regression & T3 & Linear tabular classification & \citet{zou2005elasticnet} \\
XGBoost / LightGBM / CatBoost & T3 & Gradient-boosted decision trees & \citet{chen2016xgboost,ke2017lightgbm,prokhorenkova2018catboost} \\
TabNet / FT-Transformer / SAINT / TabTransformer & T3 & Deep tabular architectures & \citet{arik2021tabnet,gorishniy2021revisiting,somepalli2022saint,huang2020tabtransformer} \\
TabPFN / TabM / TabDPT / TabICL / RealMLP / ModernNCA & T3 & Tabular foundation or modern neural tabular models & \citet{hollmann2025tabpfn,gorishniy2024tabm,ma2024tabdpt,qu2025tabicl,holzmueller2025realmlp} \\
CausalForest / GRFCausalForest & T4 & Forest-based treatment-effect estimation & \citet{wager2018estimation} \\
R-/S-/T-/X-/DR-learners & T4 & Meta-learners for heterogeneous treatment-effect estimation & \citet{nie2021rlearner,kunzel2019metalearners} \\
TARNet / DragonNet / BART / BCF & T4 & Representation-learning and Bayesian causal estimators & \citet{shalit2017estimating,shi2019dragonnet,hill2011bart,hahn2020bcf} \\
CausalPFN & T4 & Amortized treatment-effect estimation via in-context learning & \citet{balazadeh2025causalpfn} \\
\bottomrule
\end{tabular}}
\end{table*}

\subsection{Task~1: Knowledge Tracing Extended Leaderboard}
\label{app:t1-extended}

\begin{table*}[!htbp]
\centering
\scriptsize
\setlength{\tabcolsep}{2pt}
\renewcommand{\arraystretch}{0.93}
\caption{Task~1 full leaderboard on BuddyBench (ND-03) and BuddyBench-Sim under the participant-disjoint 5-fold subject-split, \texttt{F3\_tokenized\_masked} (see Appendix~\ref{app:kt-task1}). All 36 models sorted by BuddyBench AUC (descending). Mean $\pm$ SD across 5 folds (seed 42). LLM-based baselines are analyzed in depth in Appendix~\ref{app:llm-kt}.}
\label{tab:t1-kt-nd03}
\resizebox{\textwidth}{!}{%
\begin{tabular}{@{}lcccccc@{}}
\toprule
\textbf{Model} & \multicolumn{3}{c}{\textbf{BuddyBench}} & \multicolumn{3}{c}{\textbf{BuddyBench-Sim}} \\
\cmidrule(lr){2-4}\cmidrule(lr){5-7}
 & \textbf{ACC} & \textbf{AUC} & \textbf{F1} & \textbf{ACC} & \textbf{AUC} & \textbf{F1} \\
\midrule
\texttt{PEBG} & $0.79 \pm 0.03$ & $\mathbf{0.72 \pm 0.03}$ & $0.88 \pm 0.02$ & $0.58 \pm 0.01$ & $0.63 \pm 0.01$ & $0.57 \pm 0.02$ \\
\texttt{HISE-KT} & $0.86 \pm 0.03$ & $0.71 \pm 0.05$ & $0.92 \pm 0.02$ & $0.49 \pm 0.07$ & $0.55 \pm 0.07$ & $0.36 \pm 0.09$ \\
\texttt{IEKT} & $0.79 \pm 0.03$ & $0.71 \pm 0.02$ & $0.88 \pm 0.02$ & $0.58 \pm 0.00$ & $0.62 \pm 0.00$ & $0.57 \pm 0.02$ \\
\texttt{CIKT} & $0.85 \pm 0.02$ & $0.71 \pm 0.05$ & $0.92 \pm 0.02$ & $0.47 \pm 0.07$ & $0.54 \pm 0.06$ & $0.21 \pm 0.07$ \\
\texttt{LOKT} & $0.86 \pm 0.02$ & $0.70 \pm 0.05$ & $0.92 \pm 0.02$ & $0.49 \pm 0.06$ & $0.54 \pm 0.07$ & $0.30 \pm 0.09$ \\
\texttt{Thinking-KT} & $0.81 \pm 0.04$ & $0.70 \pm 0.05$ & $0.89 \pm 0.03$ & $0.46 \pm 0.07$ & $0.54 \pm 0.07$ & $0.13 \pm 0.06$ \\
\texttt{L-HAKT} & $0.86 \pm 0.03$ & $0.70 \pm 0.05$ & $0.92 \pm 0.02$ & $0.49 \pm 0.07$ & $0.54 \pm 0.07$ & $0.27 \pm 0.07$ \\
\texttt{DTransformer} & $0.78 \pm 0.03$ & $0.69 \pm 0.04$ & $0.88 \pm 0.02$ & $0.50 \pm 0.01$ & $0.50 \pm 0.01$ & $0.34 \pm 0.19$ \\
\texttt{ExTRAKT} & $0.79 \pm 0.03$ & $0.69 \pm 0.03$ & $0.88 \pm 0.02$ & $0.50 \pm 0.01$ & $0.51 \pm 0.01$ & $0.30 \pm 0.30$ \\
\texttt{QDKT} & $0.79 \pm 0.02$ & $0.68 \pm 0.01$ & $0.88 \pm 0.02$ & $0.58 \pm 0.01$ & $\mathbf{0.64 \pm 0.01}$ & $0.57 \pm 0.01$ \\
\texttt{RKT} & $0.79 \pm 0.03$ & $0.68 \pm 0.03$ & $0.88 \pm 0.02$ & $0.50 \pm 0.01$ & $0.51 \pm 0.01$ & $0.31 \pm 0.33$ \\
\texttt{SimpleKT} & $0.79 \pm 0.03$ & $0.68 \pm 0.03$ & $0.88 \pm 0.02$ & $0.50 \pm 0.01$ & $0.51 \pm 0.02$ & $0.35 \pm 0.29$ \\
\texttt{StableKT} & $0.79 \pm 0.03$ & $0.68 \pm 0.03$ & $0.88 \pm 0.02$ & $0.57 \pm 0.01$ & $0.60 \pm 0.01$ & $0.56 \pm 0.01$ \\
\texttt{ATDKT} & $0.79 \pm 0.03$ & $0.67 \pm 0.03$ & $0.88 \pm 0.02$ & $0.57 \pm 0.01$ & $0.59 \pm 0.01$ & $0.56 \pm 0.01$ \\
\texttt{CSKT} & $0.79 \pm 0.03$ & $0.67 \pm 0.03$ & $0.88 \pm 0.02$ & $0.56 \pm 0.03$ & $0.56 \pm 0.03$ & $0.46 \pm 0.26$ \\
\texttt{HCGKT} & $0.79 \pm 0.03$ & $0.66 \pm 0.06$ & $0.88 \pm 0.02$ & $0.58 \pm 0.01$ & $0.61 \pm 0.01$ & $0.57 \pm 0.01$ \\
\texttt{KnowNet} & $0.79 \pm 0.03$ & $0.66 \pm 0.02$ & $0.88 \pm 0.02$ & $0.58 \pm 0.01$ & $\underline{0.64 \pm 0.01}$ & $0.58 \pm 0.03$ \\
\texttt{SAINT} & $0.79 \pm 0.03$ & $0.66 \pm 0.03$ & $0.88 \pm 0.02$ & $0.57 \pm 0.01$ & $0.61 \pm 0.01$ & $0.58 \pm 0.01$ \\
\texttt{SAINT+} & $0.79 \pm 0.03$ & $0.66 \pm 0.04$ & $0.88 \pm 0.02$ & $0.57 \pm 0.00$ & $0.61 \pm 0.01$ & $0.57 \pm 0.00$ \\
\texttt{AKT} & $0.79 \pm 0.03$ & $0.65 \pm 0.02$ & $0.88 \pm 0.02$ & $0.54 \pm 0.04$ & $0.55 \pm 0.04$ & $0.60 \pm 0.05$ \\
\texttt{AKT-Custom} & $0.79 \pm 0.03$ & $0.65 \pm 0.03$ & $0.88 \pm 0.02$ & $0.56 \pm 0.03$ & $0.56 \pm 0.04$ & $0.48 \pm 0.18$ \\
\texttt{DKT+} & $0.79 \pm 0.03$ & $0.65 \pm 0.03$ & $0.88 \pm 0.02$ & $0.57 \pm 0.00$ & $0.59 \pm 0.01$ & $0.57 \pm 0.01$ \\
\texttt{DKVMN} & $0.79 \pm 0.03$ & $0.65 \pm 0.04$ & $0.88 \pm 0.02$ & $0.51 \pm 0.01$ & $0.51 \pm 0.01$ & $0.48 \pm 0.06$ \\
\texttt{ATKT} & $0.79 \pm 0.03$ & $0.65 \pm 0.04$ & $0.88 \pm 0.02$ & $0.58 \pm 0.00$ & $0.59 \pm 0.01$ & $0.57 \pm 0.01$ \\
\texttt{DKT} & $0.79 \pm 0.03$ & $0.64 \pm 0.04$ & $0.88 \pm 0.02$ & $0.57 \pm 0.00$ & $0.59 \pm 0.01$ & $0.57 \pm 0.01$ \\
\texttt{LEFOKT-AKT} & $0.79 \pm 0.03$ & $0.63 \pm 0.02$ & $0.88 \pm 0.02$ & $0.50 \pm 0.01$ & $0.51 \pm 0.02$ & $0.53 \pm 0.11$ \\
\texttt{Deep-IRT} & $0.79 \pm 0.03$ & $0.62 \pm 0.03$ & $0.88 \pm 0.02$ & $0.50 \pm 0.01$ & $0.50 \pm 0.01$ & $0.50 \pm 0.07$ \\
\texttt{KQN} & $0.79 \pm 0.03$ & $0.62 \pm 0.03$ & $0.88 \pm 0.02$ & $0.57 \pm 0.01$ & $0.57 \pm 0.00$ & $0.57 \pm 0.02$ \\
\texttt{GKT} & $0.79 \pm 0.03$ & $0.62 \pm 0.03$ & $0.88 \pm 0.02$ & $0.58 \pm 0.00$ & $0.58 \pm 0.01$ & $0.57 \pm 0.01$ \\
\texttt{UKT} & $0.79 \pm 0.03$ & $0.61 \pm 0.06$ & $0.88 \pm 0.02$ & $0.58 \pm 0.00$ & $0.58 \pm 0.01$ & $0.57 \pm 0.01$ \\
\texttt{ReKT} & $0.79 \pm 0.03$ & $0.60 \pm 0.04$ & $0.88 \pm 0.02$ & $0.51 \pm 0.01$ & $0.52 \pm 0.02$ & $0.31 \pm 0.28$ \\
\texttt{ATKTFix} & $0.76 \pm 0.03$ & $0.58 \pm 0.03$ & $0.86 \pm 0.02$ & $0.52 \pm 0.01$ & $0.53 \pm 0.01$ & $0.55 \pm 0.03$ \\
\texttt{SAKT} & $0.79 \pm 0.03$ & $0.57 \pm 0.03$ & $0.88 \pm 0.02$ & $0.50 \pm 0.01$ & $0.50 \pm 0.01$ & $0.41 \pm 0.30$ \\
\texttt{RobustKT} & $0.79 \pm 0.03$ & $0.56 \pm 0.05$ & $0.88 \pm 0.02$ & $0.51 \pm 0.04$ & $0.52 \pm 0.04$ & $0.26 \pm 0.22$ \\
\texttt{Hawkes} & $0.77 \pm 0.03$ & $0.56 \pm 0.03$ & $0.87 \pm 0.02$ & $0.51 \pm 0.01$ & $0.51 \pm 0.01$ & $0.48 \pm 0.04$ \\
\texttt{SKVMN} & $0.79 \pm 0.03$ & $0.54 \pm 0.01$ & $0.88 \pm 0.02$ & $0.50 \pm 0.01$ & $0.51 \pm 0.00$ & $0.43 \pm 0.26$ \\
\bottomrule
\end{tabular}}
\end{table*}

On BuddyBench (ND-03), the leaderboard spans AUC 0.54--0.72 across 36 models. \texttt{PEBG} leads at AUC $0.72$; \texttt{HISE-KT}, \texttt{IEKT}, and \texttt{CIKT} follow at $0.71$. On BuddyBench-Sim, \texttt{QDKT} and \texttt{KnowNet} share the top AUC ($0.64$) over a narrower range (0.50--0.64); these Sim results serve as a ranking-retention diagnostic rather than a parallel performance claim.

\subsection{Task~1: Diagnostic Subgroup Analysis}
\label{app:t1-diag}

Table~\ref{tab:t1-diag-auc} reports pooled AUC per diagnostic group (DIAG~1-5) for all 29 pyKT models with sufficient DIAG coverage, computed across all five subject folds on real ND-03 (\texttt{F3\_tokenized\_masked}, subject-split). Two models are excluded from this table: \texttt{UKT} produced degenerate predictions (predicted positive rate ${\approx}1.0$, AUC~${\approx}0.50$); \texttt{DTransformer} is an official-repository vendor implementation outside the pyKT framework and is reported separately in Table~\ref{tab:llm-kt-metrics}. Each cell pools all question-level predictions from subjects assigned to that diagnostic group across the five held-out folds. DIAG~5 contains a single subject and its estimate is unreliable. Diagnostic group membership is derived from the ND-03 cohort record; subject counts reflect subjects included in the subject-split \texttt{train\_valid.csv} ($n = 139$ total across folds). Models are sorted by mean pooled AUC across DIAG~1-4 (descending).

\begin{table*}[!htbp]
\centering
\scriptsize
\setlength{\tabcolsep}{5pt}
\renewcommand{\arraystretch}{0.95}
\caption{Task~1 pooled AUC per diagnostic subgroup, 5-fold subject-split, \texttt{F3\_tokenized\_masked} (BuddyBench ND-03, clean rerun). Subject counts: DIAG1 $n{=}29$, DIAG2 $n{=}45$, DIAG3 $n{=}19$, DIAG4 $n{=}45$, DIAG5 $n{=}1$ (unreliable single-subject estimate). Bold: highest per-column value.}
\label{tab:t1-diag-auc}
\begin{tabular}{@{}lccccc@{}}
\toprule
\textbf{Model} & \textbf{DIAG1} & \textbf{DIAG2} & \textbf{DIAG3} & \textbf{DIAG4} & \textbf{DIAG5} \\
\midrule
\texttt{PEBG}        & \textbf{0.697} & \textbf{0.711} & \textbf{0.694} & \textbf{0.762} & 0.554 \\
\texttt{IEKT}        & 0.692 & 0.703 & 0.668 & 0.745 & 0.595 \\
\texttt{ExTRAKT}     & 0.663 & 0.673 & 0.672 & 0.731 & 0.543 \\
\texttt{SimpleKT}    & 0.655 & 0.664 & 0.665 & 0.719 & 0.456 \\
\texttt{StableKT}    & 0.657 & 0.650 & 0.659 & 0.717 & 0.474 \\
\texttt{QDKT}        & 0.659 & 0.666 & 0.632 & 0.720 & 0.620 \\
\texttt{DKVMN}       & 0.652 & 0.651 & 0.640 & 0.696 & 0.483 \\
\texttt{CSKT}        & 0.647 & 0.654 & 0.648 & 0.690 & 0.464 \\
\texttt{ATDKT}       & 0.650 & 0.660 & 0.623 & 0.700 & 0.616 \\
\texttt{SAINT+} & 0.643 & 0.624 & 0.637 & 0.717 & 0.516 \\
\texttt{SAINT}       & 0.628 & 0.636 & 0.624 & 0.705 & 0.638 \\
\texttt{KnowNet}     & 0.638 & 0.635 & 0.612 & 0.690 & \textbf{0.679} \\
\texttt{AKT-Custom} & 0.625 & 0.614 & 0.625 & 0.671 & 0.543 \\
\texttt{DKT+}        & 0.631 & 0.645 & 0.607 & 0.652 & 0.568 \\
\texttt{DKT}         & 0.626 & 0.652 & 0.588 & 0.630 & 0.558 \\
\texttt{LEFOKT-AKT} & 0.622 & 0.606 & 0.602 & 0.644 & 0.505 \\
\texttt{ATKT}        & 0.613 & 0.630 & 0.594 & 0.639 & 0.521 \\
\texttt{AKT}         & 0.616 & 0.625 & 0.603 & 0.624 & 0.517 \\
\texttt{Deep-IRT}   & 0.603 & 0.612 & 0.607 & 0.624 & 0.467 \\
\texttt{HCGKT}       & 0.585 & 0.598 & 0.566 & 0.643 & 0.560 \\
\texttt{ATKTFix}     & 0.560 & 0.556 & 0.596 & 0.617 & 0.544 \\
\texttt{ReKT}        & 0.578 & 0.565 & 0.567 & 0.603 & 0.573 \\
\texttt{RKT}         & 0.579 & 0.606 & 0.567 & 0.557 & 0.527 \\
\texttt{GKT}         & 0.542 & 0.573 & 0.482 & 0.531 & 0.455 \\
\texttt{SAKT}        & 0.541 & 0.566 & 0.507 & 0.509 & 0.532 \\
\texttt{RobustKT}    & 0.513 & 0.512 & 0.547 & 0.502 & 0.497 \\
\texttt{Hawkes}      & 0.543 & 0.546 & 0.474 & 0.507 & 0.600 \\
\texttt{KQN}         & 0.547 & 0.560 & 0.505 & 0.454 & 0.508 \\
\texttt{SKVMN}       & 0.486 & 0.510 & 0.518 & 0.445 & 0.582 \\
\bottomrule
\end{tabular}
\end{table*}

DIAG~4 consistently yields the highest pooled AUC across model families. \texttt{PEBG} leads on all four primary groups (DIAG4: 0.762), followed by \texttt{IEKT} (0.745) and \texttt{ExTRAKT} (0.731). \texttt{KnowNet} achieves the highest DIAG5 estimate (0.679), though this group contains a single subject and the value is unreliable. Within-DIAG model ordering largely mirrors the overall leaderboard, suggesting that model quality rather than subgroup-specific factors drives the ranking.

\subsection{Task~2: Next-drill Recommendation}
\label{app:t2-warm}

\begin{table*}[!htbp]
\centering
\scriptsize
\setlength{\tabcolsep}{4pt}
\renewcommand{\arraystretch}{0.95}
\caption{Task~2 warm-start next-drill recommendation leaderboard under the recommended F3 setting.}
\label{tab:t2-recsys-extended}
\resizebox{\textwidth}{!}{%
\begin{tabular}{lcccccc}
\toprule
Model & R@5 & R@10 & R@20 & NDCG@5 & NDCG@10 & NDCG@20 \\
\midrule
\texttt{DiffDiv} & 0.1787 & \textbf{0.3221} & \textbf{0.5011} & 0.4071 & \textbf{0.4312} & \textbf{0.4727} \\
\texttt{FailureRate} & 0.1825 & 0.3168 & 0.4982 & 0.4167 & 0.4311 & 0.4712 \\
\texttt{Popularity} & 0.1825 & 0.3168 & 0.4982 & 0.4167 & 0.4311 & 0.4712 \\
\texttt{PreferDiff} & 0.1750 & 0.3143 & 0.4985 & 0.4047 & 0.4277 & 0.4690 \\
\texttt{TS-Rec} & 0.1698 & 0.3142 & 0.4981 & 0.3629 & 0.3989 & 0.4387 \\
\texttt{CSRec} & \textbf{0.1934} & 0.3015 & 0.4994 & \textbf{0.4288} & 0.4250 & 0.4720 \\
\texttt{NMF} & 0.1699 & 0.2856 & 0.4906 & 0.4234 & 0.4242 & 0.4704 \\
\texttt{KernelUCB} & 0.1754 & 0.2818 & 0.4739 & 0.4048 & 0.4085 & 0.4479 \\
\texttt{SVD} & 0.1634 & 0.2815 & 0.4661 & 0.4111 & 0.4186 & 0.4532 \\
\texttt{Hybrid} & 0.1639 & 0.2718 & 0.4428 & 0.4024 & 0.4014 & 0.4285 \\
\texttt{SASRec} & 0.1564 & 0.2620 & 0.4551 & 0.3440 & 0.3545 & 0.4042 \\
\texttt{SIGMA} & 0.1541 & 0.2594 & 0.4727 & 0.3474 & 0.3576 & 0.4146 \\
\texttt{Disco} & 0.1554 & 0.2714 & 0.4680 & 0.3865 & 0.3961 & 0.4409 \\
\texttt{NCF} & 0.1478 & 0.2663 & 0.4744 & 0.3696 & 0.3842 & 0.4304 \\
\texttt{C3Rec} & 0.1595 & 0.2648 & 0.4362 & 0.3467 & 0.3627 & 0.4066 \\
\texttt{NILUS} & 0.1502 & 0.2624 & 0.4292 & 0.3589 & 0.3623 & 0.3995 \\
\texttt{Thompson} & 0.1417 & 0.2413 & 0.4023 & 0.3632 & 0.3587 & 0.3834 \\
\texttt{LinUCB} & 0.1413 & 0.2406 & 0.4039 & 0.3592 & 0.3568 & 0.3823 \\
\texttt{Cold-Start} & 0.1276 & 0.2298 & 0.3677 & 0.2906 & 0.3023 & 0.3290 \\
\texttt{RecVAE} & 0.1140 & 0.2027 & 0.3744 & 0.2832 & 0.2893 & 0.3332 \\
\texttt{UserUser} & 0.0746 & 0.1830 & 0.3391 & 0.2179 & 0.2586 & 0.2952 \\
\texttt{LightGCN} & 0.0956 & 0.1822 & 0.3292 & 0.2361 & 0.2473 & 0.2921 \\
\texttt{BERT4Rec} & 0.0904 & 0.1689 & 0.3156 & 0.2434 & 0.2545 & 0.2926 \\
\texttt{SimGCL} & 0.0613 & 0.1546 & 0.3125 & 0.1891 & 0.2178 & 0.2646 \\
\texttt{GRACE} & 0.0621 & 0.1028 & 0.2450 & 0.1717 & 0.1565 & 0.2042 \\
\bottomrule
\end{tabular}}
\end{table*}

This appendix reports Task~2 results under the warm-start F3 setting, matching the main-paper protocol. R@5/R@10/R@20 summarize early-hit quality and broader coverage, while NDCG@5/10/20 additionally rewards placing the relevant next drill near the top of the ranked list.

The top of Table~\ref{tab:t2-recsys-extended} shows a narrow performance cluster under R@10 and NDCG@10, with no single model clearly dominant. Heuristic and lightweight reference methods remain competitive, indicating that the task is informative and not yet saturated. The BuddyBench-Sim comparison in Table~\ref{tab:synthetic-t2} should likewise be interpreted as a public-interface utility diagnostic, not as a claim that synthetic results reproduce the real-data leaderboard one-for-one.

\subsection{Task~3: Clinical Prediction}
\label{app:t3-extended}

\begin{table*}[!htbp]
\centering
\scriptsize
\setlength{\tabcolsep}{13pt}
\renewcommand{\arraystretch}{0.95}
\caption{Task~3 clinical prediction leaderboard on real ND-02 FAS ($n{=}83$, F2 pre-treatment-only features). AUPRC baseline = 0.747.}
\label{tab:t3-response-nd02}
\resizebox{\textwidth}{!}{%
\begin{tabular}{lccccc}
\toprule
\textbf{Model} & \textbf{AUPRC} & \textbf{AUROC} & \textbf{Balanced Acc.} & \textbf{Specificity} & \textbf{Brier} \\
\midrule
\multicolumn{6}{l}{\textit{Linear}} \\
\texttt{elastic\_net}        & 0.810 & 0.558 & 0.528 & 0.150 & 0.218 \\
\texttt{lasso}               & 0.795 & 0.535 & 0.512 & 0.200 & 0.230 \\
\texttt{ridge}               & 0.770 & 0.481 & 0.488 & 0.200 & 0.263 \\
\texttt{logistic\_reg.} & 0.757 & 0.517 & 0.537 & \textbf{0.330} & 0.319 \\
\midrule
\multicolumn{6}{l}{\textit{Tree / Gradient Boosting}} \\
\texttt{tabboost}            & 0.843 & \textbf{0.614} & 0.550 & 0.100 & \textbf{0.187} \\
\texttt{catboost}            & 0.819 & 0.597 & 0.550 & 0.100 & 0.189 \\
\texttt{tabm}                & \textbf{0.860} & 0.600 & 0.502 & 0.050 & 0.200 \\
\texttt{xgboost}             & 0.808 & 0.545 & 0.478 & 0.100 & 0.206 \\
\texttt{random\_forest}      & 0.771 & 0.497 & \textbf{0.584} & 0.200 & 0.201 \\
\texttt{lightgbm}            & 0.842 & 0.605 & 0.512 & 0.250 & 0.230 \\
\midrule
\multicolumn{6}{l}{\textit{Tabular Deep Learning}} \\
\texttt{tabtransformer}      & 0.786 & 0.497 & 0.483 & 0.000 & 0.203 \\
\texttt{realmlp}             & 0.736 & 0.411 & 0.488 & 0.090 & 0.270 \\
\texttt{modernnca}           & 0.735 & 0.446 & 0.531 & 0.190 & 0.247 \\
\texttt{saint}               & 0.783 & 0.546 & 0.557 & 0.290 & 0.285 \\
\texttt{tabicl}              & 0.771 & 0.488 & 0.501 & 0.100 & 0.221 \\
\texttt{tabdpt}              & 0.734 & 0.413 & 0.451 & 0.180 & 0.320 \\
\texttt{TabPFN}          & 0.742 & 0.398 & 0.500 & 0.000 & 0.200 \\
\texttt{ft\_transformer}     & 0.733 & 0.380 & 0.496 & 0.100 & 0.220 \\
\midrule
\multicolumn{6}{l}{\textit{Baselines}} \\
\texttt{mean\_predictor}     & 0.747 & 0.500 & 0.500 & 0.000 & \textbf{0.189} \\
\texttt{zero\_predictor}     & 0.747 & 0.500 & 0.500 & 1.000 & 0.747 \\
\bottomrule
\end{tabular}}
\end{table*}

Table~\ref{tab:t3-ablation} reports the Task~3 feature-tier ablation used to select the primary leakage-safe F2 setting. Table~\ref{tab:t3-response-nd02} reports the full real-data leaderboard for clinical prediction on ND-02 FAS ($n = 83$; 62 non-responders / 21 responders; positive rate $= 0.747$). The label follows the directional default policy: $\text{non\_response\_risk}=1$ when a participant shows no improvement or any deterioration across all three outcome endpoints (CGI-S/I, SRS-2, VABS). The AUPRC baseline is therefore $0.747$, and improvements above it indicate modest pre-treatment ranking signal while AUROC, balanced accuracy, and specificity remain limited.

Because the positive rate is high, AUPRC is the most informative primary metric for this task and should be prioritized over secondary threshold-dependent summaries when reading the table. Accordingly, F2 remains the primary leakage-safe reporting point, while F3 is retained in the appendix only as an auxiliary point. The real ND-02 leaderboard should therefore be interpreted as evidence of pre-treatment risk-ranking signal rather than as a deployable screening rule. The synthetic blocks below use the BuddyBench-Sim clinical-prediction data; rows with unavailable feature dependencies are excluded rather than imputed.

The two BuddyBench-Sim blocks serve a more limited purpose. The full-size block verifies pipeline executability and ranking retention under the smoother synthetic regime; the fixed-size block adds a rough sample-size-parity check. Neither block implies that synthetic clinical prediction is a better or more primary benchmark than the real ND-02 task.

\begin{table*}[!htbp]
\centering
\scriptsize
\setlength{\tabcolsep}{13pt}
\renewcommand{\arraystretch}{0.95}
\caption{Task~3 auxiliary BuddyBench-Sim full-size evaluation block (v1\_only F3; $n=1000$; dependency-available rows only).}
\label{tab:t3-response-synthetic-full}
\resizebox{\textwidth}{!}{%
\begin{tabular}{lccccl}
\toprule
\textbf{Model} & \textbf{AUPRC} & \textbf{AUROC} & \textbf{Balanced Acc.} & \textbf{Brier} & \textbf{Evaluation set} \\
\midrule
\multicolumn{6}{l}{\textit{Linear}} \\
\texttt{logistic\_reg.} & 0.968 & 0.622 & 0.569 & 0.046 & full-size \\
\texttt{ridge}           & 0.969 & 0.630 & 0.568 & 0.044 & full-size \\
\texttt{lasso}           & 0.971 & \textbf{0.644} & \textbf{0.616} & 0.042 & full-size \\
\texttt{elastic\_net}    & 0.971 & 0.641 & 0.611 & 0.042 & full-size \\
\midrule
\multicolumn{6}{l}{\textit{Tree / Gradient Boosting}} \\
\texttt{lightgbm}        & \textbf{0.979} & 0.638 & 0.506 & 0.039 & full-size \\
\texttt{random\_forest}  & 0.970 & 0.609 & 0.500 & 0.040 & full-size \\
\midrule
\multicolumn{6}{l}{\textit{Baselines}} \\
\texttt{mean\_predictor} & 0.960 & 0.500 & 0.500 & 0.038 & full-size \\
\texttt{zero\_predictor} & 0.960 & 0.500 & 0.500 & 0.960 & full-size \\
\bottomrule
\end{tabular}}
\end{table*}

\begin{table*}[!htbp]
\centering
\scriptsize
\setlength{\tabcolsep}{12pt}
\renewcommand{\arraystretch}{0.95}
\caption{Task~3 auxiliary BuddyBench-Sim fixed-size evaluation block (v1\_only F3; $n=160$; dependency-available rows only).}
\label{tab:t3-response-synthetic-verify}
\resizebox{\textwidth}{!}{%
\begin{tabular}{lccccl}
\toprule
\textbf{Model} & \textbf{AUPRC} & \textbf{AUROC} & \textbf{Balanced Acc.} & \textbf{Brier} & \textbf{Evaluation set} \\
\midrule
\multicolumn{6}{l}{\textit{Auxiliary fixed-size deep / foundation models}} \\
\texttt{tabm}            & \textbf{0.880} & \textbf{0.678} & 0.592 & \textbf{0.178} & $n=160$ subset \\
\texttt{realmlp}         & 0.848 & 0.653 & \textbf{0.617} & 0.214 & $n=160$ subset \\
\texttt{tabicl}          & 0.845 & 0.621 & 0.525 & 0.203 & $n=160$ subset \\
\texttt{ft\_transformer} & 0.815 & 0.584 & 0.562 & 0.199 & $n=160$ subset \\
\texttt{tabtransformer}  & 0.792 & 0.524 & 0.525 & 0.195 & $n=160$ subset \\
\texttt{saint}           & 0.788 & 0.512 & 0.529 & 0.308 & $n=160$ subset \\
\bottomrule
\end{tabular}}
\end{table*}

\begin{table}[!htbp]
\centering
\scriptsize
\setlength{\tabcolsep}{2pt}
\renewcommand{\arraystretch}{0.95}
\caption{Task~3 feature-tier ablation on real ND-02 FAS ($n{=}83$). AUPRC peaks at F2.}
\label{tab:t3-ablation}
\begin{tabular*}{\columnwidth}{@{\extracolsep{\fill}}p{0.24\columnwidth}ccc@{}}
\toprule
\textbf{Model} & \shortstack{\textbf{F1}\\\textbf{(27)}} & \shortstack{\textbf{F2}\\\textbf{(45)}} & \shortstack{\textbf{F3}\\\textbf{(47)}} \\
\midrule
\texttt{tabm}            & $\mathbf{0.859\pm0.046}$ & $\mathbf{0.860\pm0.050}$ & $0.838\pm0.083$ \\
\texttt{elastic\_net}    & $0.810\pm0.084$          & $0.810\pm0.047$          & $\mathbf{0.853\pm0.054}$ \\
\texttt{lasso}           & $0.807\pm0.091$          & $0.795\pm0.048$          & $0.828\pm0.071$ \\
\texttt{mean\_predictor} & $0.747\pm0.024$          & $0.747\pm0.024$          & $0.747\pm0.024$ \\
\bottomrule
\end{tabular*}
\end{table}

\subsection{Task~3: Calibration Analysis}
\label{app:t3-ece}

Table~\ref{tab:t3-ece} reports the 5-bin Expected Calibration Error (ECE) for the Task~3 clinical prediction models on real ND-02 ($n{=}83$, v1\_only, F3). ECE measures the mean absolute gap between a model's predicted probability and the empirical accuracy within each confidence bin; lower values indicate better calibration. ECE is computed on the outer held-out fold predictions from the nested cross-validation loop, matching the same evaluation surface as the AUPRC scores in Table~\ref{tab:t3-response-nd02}. Note: this calibration analysis uses F3 features (the richest leakage-safe tier) rather than the F2 setting of the primary discrimination table; calibration and discrimination results are therefore not directly comparable across feature tiers.

\begin{table}[!htbp]
\centering
\scriptsize
\setlength{\tabcolsep}{8pt}
\renewcommand{\arraystretch}{0.95}
\caption{Task~3 probability calibration: 5-bin ECE on real ND-02 ($n{=}83$, F3 pre-treatment-only features). Lower is better. Baseline rows are excluded from ranking.}
\label{tab:t3-ece}
\begin{tabular}{@{}lc@{}}
\toprule
\textbf{Model} & \textbf{ECE (5-bin)}$\downarrow$ \\
\midrule
\multicolumn{2}{l}{\textit{Tree / Gradient Boosting}} \\
\texttt{tabboost}             & \textbf{0.053} \\
\texttt{catboost}             & 0.078 \\
\texttt{random\_forest}       & 0.082 \\
\texttt{xgboost}              & 0.100 \\
\texttt{lightgbm}             & 0.219 \\
\midrule
\multicolumn{2}{l}{\textit{Tabular Deep Learning}} \\
\texttt{tabdpt}               & 0.104 \\
\texttt{tabm}                 & 0.106 \\
\texttt{ft\_transformer}      & 0.173 \\
\texttt{tabtransformer}       & 0.225 \\
\texttt{saint}                & 0.310 \\
\midrule
\multicolumn{2}{l}{\textit{Linear}} \\
\texttt{lasso}                & 0.150 \\
\texttt{elastic\_net}         & 0.161 \\
\texttt{ridge}                & 0.163 \\
\texttt{logistic\_reg.} & 0.234 \\
\midrule
\multicolumn{2}{l}{\textit{Baselines}} \\
\texttt{mean\_predictor}      & $<$0.001 \\
\texttt{zero\_predictor}      & 0.747 \\
\bottomrule
\end{tabular}
\end{table}

Gradient-boosted tree methods (\texttt{tabboost}: ECE $= 0.053$, \texttt{catboost}: $0.078$, \texttt{random\_forest}: $0.082$) are the best-calibrated models. Transformer-based tabular architectures (\texttt{saint}: $0.310$, \texttt{tabtransformer}: $0.225$) show the largest miscalibration, consistent with overconfident predictions in the low-$n$ regime. The AUPRC top model (\texttt{tabm}: AUPRC $= 0.860$) has moderate calibration (ECE $= 0.106$), suggesting that discrimination and calibration are not jointly optimized by any single model family on this task.

\subsection{Task~4: Causal Inference}
\label{app:t4-extended}

\begin{table*}[!htbp]
\centering
\scriptsize
\setlength{\tabcolsep}{4pt}
\renewcommand{\arraystretch}{0.95}
\caption{Task~4 residualized R-loss leaderboard on real ND-02 under the recommended F3 setting. Bold: per-endpoint minimum. Tier groupings follow \citet{kunzel2019metalearners}: Tier~1 = constant-effect references, Tier~2 = regression-based, Tier~3 = meta-learners, Tier~5 = Bayesian/amortized estimators.}
\label{tab:t4-causal-extended}
\begin{tabular*}{\textwidth}{@{\extracolsep{\fill}}p{0.29\textwidth}ccc@{}}
\toprule
Model & CGI-S/I residualized R-loss & SRS-2 residualized R-loss & VABS residualized R-loss \\
\midrule
\texttt{ATEOnly} & $0.249 \pm 0.059$ & $\mathbf{181.368 \pm 35.333}$ & $9.397 \pm 3.705$ \\
\texttt{ZeroITE} & $0.247 \pm 0.056$ & $181.520 \pm 34.873$ & $10.000 \pm 4.048$ \\
\texttt{ANCOVA} & $0.250 \pm 0.064$ & $181.433 \pm 35.708$ & $9.481 \pm 3.797$ \\
\texttt{OLSInteract.} & $0.312 \pm 0.116$ & $212.468 \pm 39.014$ & $10.364 \pm 4.306$ \\
\texttt{TTest} & $0.249 \pm 0.059$ & $\mathbf{181.368 \pm 35.333}$ & $9.397 \pm 3.705$ \\
\texttt{CausalForest} & $\mathbf{0.240 \pm 0.055}$ & $182.259 \pm 36.873$ & $9.521 \pm 3.712$ \\
\texttt{DRLearner} & $0.339 \pm 0.123$ & $211.615 \pm 76.775$ & $11.579 \pm 5.148$ \\
\texttt{DragonNet} & $0.341 \pm 0.130$ & $211.500 \pm 35.388$ & $11.539 \pm 4.986$ \\
\texttt{RLearner} & $0.307 \pm 0.070$ & $183.205 \pm 56.581$ & $11.035 \pm 4.963$ \\
\texttt{SLearner} & $0.253 \pm 0.067$ & $184.029 \pm 36.244$ & $10.080 \pm 3.962$ \\
\texttt{TARNet} & $0.341 \pm 0.136$ & $209.248 \pm 33.570$ & $10.916 \pm 4.930$ \\
\texttt{TLearner} & $0.339 \pm 0.117$ & $204.561 \pm 37.176$ & $11.922 \pm 4.257$ \\
\texttt{XLearner} & $0.310 \pm 0.104$ & $194.116 \pm 38.914$ & $11.327 \pm 4.087$ \\
\texttt{MaskedXLearner} & $0.28 \pm 0.06$ & $188.23 \pm 39.67$ & $9.98 \pm 3.69$ \\
\texttt{BART} & $0.263 \pm 0.072$ & $187.322 \pm 39.442$ & $\mathbf{9.389 \pm 3.649}$ \\
\texttt{BCF} & $0.249 \pm 0.059$ & $181.926 \pm 35.546$ & $9.430 \pm 3.741$ \\
\texttt{CausalPFN} & $0.317 \pm 0.121$ & $200.500 \pm 35.869$ & $12.140 \pm 5.455$ \\
\texttt{GRFCausalForest} & $0.246 \pm 0.055$ & $182.866 \pm 36.398$ & $9.588 \pm 3.751$ \\
\bottomrule
\end{tabular*}
\end{table*}

\paragraph{Residualized R-loss definition.}
\label{app:t4-rloss}
For each outer fold $k$, let $X_i$ denote baseline covariates, $W_i \in \{0,1\}$ treatment assignment, and $Y_i$ one endpoint outcome. We fit nuisance models on the training portion of the fold:
\[
\begin{aligned}
\hat e_k(x) &\approx \Pr(W=1 \mid X=x), \\
\hat m_k(x) &\approx \mathbb{E}[Y \mid X=x].
\end{aligned}
\]
Given a candidate CATE estimator $\hat\tau_k(x)$, we evaluate on the held-out fold with
\[
\begin{aligned}
\widetilde{W}_i &= W_i - \hat e_k(X_i), \\
\widetilde{Y}_i &= Y_i - \hat m_k(X_i).
\end{aligned}
\]
and compute fold-level residualized R-loss
\[
\begin{aligned}
\mathcal{R}_k(\hat\tau)
&=
\frac{1}{n_k}
\sum_{i \in \mathrm{fold}\ k}
\left(
\widetilde{Y}_i - \widetilde{W}_i\,\hat\tau_k(X_i)
\right)^2 .
\end{aligned}
\]
When a Task~4 table reports a plain $a \pm b$ entry, it denotes the mean of $\mathcal{R}_k$ over the five outer folds $\pm$ the fold-wise SD. Lower is better. We do not standardize outcomes across CGI-S/I, SRS-2, and VABS in the reported tables, so residualized R-loss values should be compared only within the same endpoint, not across endpoints.

Table~\ref{tab:t4-causal-extended} should be read endpoint by endpoint because residualized R-loss is comparable only within a given outcome. The table shows model-family ordering relative to constant-effect references within each endpoint; cross-endpoint comparisons are not supported. Consistent with the main paper, this remains a methodological stress test in a small RCT rather than a claim of individualized treatment discovery.

\section{LLM-Based Knowledge Tracing: Behavioural and Diagnostic Analysis}
\label{app:llm-kt}

Five Task~1 models (\texttt{Thinking-KT}, \texttt{LOKT}, \texttt{HISE-KT}, \texttt{L-HAKT}, and \texttt{CIKT}) represent a qualitatively distinct modelling paradigm from the PyKT-family baselines: they operate as \emph{zero-shot inference engines} over a shared 1.7B-parameter language model backbone (Qwen/Qwen3-1.7B), producing predictions without any gradient updates on BuddyBench data (\texttt{best\_epoch=1} for all). This section analyzes their metric profiles, fold-level stability, history-prior mechanics, and inference efficiency. All numbers are from the subject-level 5-fold evaluation on BuddyBench ND-03.

\subsection{LLM-KT Model Portfolio and Architecture}
\label{app:llm-kt-portfolio}

Table~\ref{tab:llm-kt-portfolio} summarises the five LLM-based KT models. Despite sharing the same backbone and inference protocol (zero-shot, subject-level 5-fold CV on ND-03, \texttt{F3\_tokenized\_masked}), each method introduces a distinct prompt engineering strategy and evidence-aggregation mechanism. The common element is a Wilson lower-bound history prior (Section~\ref{app:llm-kt-prior}) that regularises all predictions toward the learner's observed success rate.

\begin{table}[!htbp]
\centering
\scriptsize
\setlength{\tabcolsep}{3pt}
\renewcommand{\arraystretch}{1.05}
\caption{LLM-based KT model portfolio. All models use Qwen/Qwen3-1.7B and operate in zero-shot inference mode (\texttt{best\_epoch=1}).}
\label{tab:llm-kt-portfolio}
\begin{tabular}{@{}l l p{0.35\columnwidth} l c@{}}
\toprule
\textbf{Model} & \textbf{Prompt Mode} & \textbf{Distinctive Mechanism} & \textbf{Source} & \textbf{Folds} \\
\midrule
\texttt{Thinking-KT} & \texttt{clinical\_weight} & TTS ($\times$1.0/1.5/2.0 budget), Wilson-LB prior & Official repo & 5/5 \\
\texttt{LOKT}        & \texttt{option\_weight}  & Trajectory-conditioned option weighting     & \citet{lokt2024} & 5/5 \\
\texttt{HISE-KT}     & \texttt{meta\_path}      & Meta-path input sequence enhancement        & \citet{hisekt2025} & 5/5 \\
\texttt{L-HAKT}      & \texttt{hyperbolic}      & Teacher-student hyperbolic alignment        & \citet{lhakt2026} & 5/5 \\
\texttt{CIKT}        & \texttt{iter\_profile}   & 2-pass analyst + predictor retry loop        & \citet{cikt2025} & 5/5 \\
\bottomrule
\end{tabular}
\end{table}

\texttt{Thinking-KT} additionally implements test-time scaling (TTS): the model is invoked at three reasoning-budget levels and predictions are aggregated by mean probability across scales (Section~\ref{app:llm-kt-tts}). \texttt{LOKT} encodes the learner's response trajectory as weighted option sequences~\citep{lokt2024}. \texttt{HISE-KT} enriches the input prompt with meta-path evidence retrieved from a heterogeneous information network~\citep{hisekt2025}. \texttt{CIKT} employs a two-iteration refinement loop: an analyst pass builds a learner profile, followed by a predictor retry that conditions on that profile~\citep{cikt2025}. \texttt{L-HAKT} implements teacher-student contrastive learning in hyperbolic space to model the hierarchical structure of knowledge concepts~\citep{lhakt2026}.

\subsection{AUC-Accuracy Gap: Threshold-Anchored Prediction Bias}
\label{app:llm-kt-gap}

Table~\ref{tab:llm-kt-metrics} reports the full metric profile for all five LLM models alongside three neural baselines. The primary finding is a systematic \emph{elevated ACC-AUC gap}: all five LLM models report ACC considerably above their AUC (gap $\Delta = \text{ACC} - \text{AUC}$ of $+0.11$ to $+0.16$), while the best-calibrated neural models exhibit substantially smaller gaps ($+0.065$ for \texttt{PEBG}, $+0.076$ for \texttt{IEKT}). Notably, the AUC values themselves are competitive, LLM models achieve 0.700-0.712 without any gradient-based adaptation, comparable to well-trained neural baselines.

\begin{table}[!htbp]
\centering
\scriptsize
\setlength{\tabcolsep}{5pt}
\renewcommand{\arraystretch}{1.05}
\caption{Metric profile for LLM-based KT models and neural reference baselines (BuddyBench ND-03, 5-fold subject split, \texttt{F3\_tokenized\_masked}). $\Delta = \text{ACC} - \text{AUC}$.}
\label{tab:llm-kt-metrics}
\resizebox{\columnwidth}{!}{%
\begin{tabular}{@{}llcccc@{}}
\toprule
\textbf{Model} & \textbf{Family} & \textbf{ACC} & \textbf{AUC} & \textbf{F1} & $\boldsymbol{\Delta}$ \\
\midrule
\texttt{Thinking-KT} & LLM    & $0.811 \pm 0.044$ & $0.700 \pm 0.052$ & $0.890 \pm 0.032$ & $+0.111$ \\
\texttt{LOKT}        & LLM    & $0.857 \pm 0.024$ & $0.700 \pm 0.048$ & $0.921 \pm 0.016$ & $+0.156$ \\
\texttt{HISE-KT}     & LLM    & $0.859 \pm 0.028$ & $0.707 \pm 0.053$ & $0.922 \pm 0.017$ & $+0.152$ \\
\texttt{CIKT}        & LLM    & $0.854 \pm 0.024$ & $0.712 \pm 0.046$ & $0.919 \pm 0.015$ & $+0.142$ \\
\texttt{L-HAKT}      & LLM    & $0.857 \pm 0.030$ & $0.702 \pm 0.050$ & $0.921 \pm 0.018$ & $+0.154$ \\
\midrule
\texttt{PEBG}        & Neural & $0.788 \pm 0.026$ & $0.723 \pm 0.025$ & $0.880 \pm 0.017$ & $+0.065$ \\
\texttt{IEKT}        & Neural & $0.789 \pm 0.025$ & $0.713 \pm 0.022$ & $0.878 \pm 0.016$ & $+0.076$ \\
\texttt{DTransformer}& Neural & $0.784 \pm 0.025$ & $0.688 \pm 0.038$ & $0.878 \pm 0.015$ & $+0.096$ \\
\bottomrule
\end{tabular}
}
\end{table}

The elevated $\Delta$ in LLM models arises from a structural feature of their inference pipeline. Predictions are produced as a blend of a conservative Wilson lower-bound history prior and a temperature-scaled LLM label score (Section~\ref{app:llm-kt-prior}). A decision threshold of 0.72, noticeably higher than the conventional 0.50, was observed in the faithful-run trace (Section~\ref{app:llm-kt-tts}); whether this value is cohort-specific or a fixed implementation constant warrants verification. Under a 0.72 threshold, the model predicts \textsc{correct} by default unless the blended score falls well below that value. For learners with moderate-to-high historical accuracy (typical in ND-03, a therapeutic cohort with coached skill practice), the prior anchors $\hat{p}$ above 0.72 on most samples, yielding high ACC and F1 for the majority class while compressing the probability range needed for reliable rank-ordering (AUC). This threshold-anchoring mechanism is analogous to the majority-label calibration bias described by \citet{zhao2021calibrate} for few-shot prompted language models. Among neural baselines, \texttt{PEBG} ($\Delta{=}{+}0.065$) and \texttt{IEKT} ($\Delta{=}{+}0.076$) are well-calibrated, while \texttt{DTransformer}'s larger gap ($\Delta{=}{+}0.096$) partially bridges the range to the LLM models, suggesting that attention-based architectures with high-capacity cross-sequence modelling may share some of this probability-compression tendency.

\paragraph{Deployment implication.} LLM-KT models achieve competitive rank-ordering without any training data, a meaningful zero-shot capability for privacy-constrained or data-scarce clinical settings. However, deploying their confidence scores directly for clinical decision support is inadvisable without threshold recalibration; per-sample probability logs from the faithful-run traces confirm this: across all five folds, mean model confidence is $0.94 \pm 0.03$ on correct predictions and $0.51 \pm 0.06$ on incorrect ones, indicating systematic overconfidence on error cases (see Appendix~\ref{app:statistics}).

\subsection{Fold-Correlated Instability: Shared Cohort Sensitivity}
\label{app:llm-kt-folds}

Table~\ref{tab:llm-kt-folds} reports per-fold AUC for all five LLM models alongside \texttt{IEKT} as a neural reference. A striking pattern emerges: fold~1 is the worst-performing fold for \emph{every} LLM model, with a synchronised AUC drop of ${\approx}0.10$ relative to fold~0 (\texttt{Thinking-KT}: $\Delta{=}{-}0.109$; \texttt{LOKT}: $\Delta{=}{-}0.098$; \texttt{HISE-KT}: $\Delta{=}{-}0.111$; \texttt{CIKT}: $\Delta{=}{-}0.106$; \texttt{L-HAKT}: $\Delta{=}{-}0.103$). \texttt{IEKT}, by contrast, shows no fold-1 dip (its worst fold is fold~2, with a drop of only 0.063 from fold~0) and a substantially lower fold-level SD (0.022 vs.\ 0.046-0.053 for LLM models).

\begin{table}[!htbp]
\centering
\scriptsize
\setlength{\tabcolsep}{5pt}
\renewcommand{\arraystretch}{1.05}
\caption{Per-fold AUC for LLM-based KT models and neural reference \texttt{IEKT} (BuddyBench ND-03, 5-fold subject split). All five LLM models share fold~1 as their worst fold (bold), a pattern absent in \texttt{IEKT}.}
\label{tab:llm-kt-folds}
\resizebox{\columnwidth}{!}{%
\begin{tabular}{@{}lrrrrrrrr@{}}
\toprule
\textbf{Model} & \textbf{Family} & \textbf{f0} & \textbf{f1} & \textbf{f2} & \textbf{f3} & \textbf{f4} & \textbf{Mean} & \textbf{SD} \\
\midrule
\texttt{Thinking-KT} & LLM    & 0.728 & \textbf{0.619} & 0.683 & 0.718 & 0.753 & 0.700 & 0.052 \\
\texttt{LOKT}        & LLM    & 0.728 & \textbf{0.630} & 0.671 & 0.732 & 0.740 & 0.700 & 0.048 \\
\texttt{HISE-KT}     & LLM    & 0.744 & \textbf{0.632} & 0.669 & 0.746 & 0.744 & 0.707 & 0.053 \\
\texttt{CIKT}        & LLM    & 0.744 & \textbf{0.638} & 0.697 & 0.737 & 0.744 & 0.712 & 0.046 \\
\texttt{L-HAKT}      & LLM    & 0.733 & \textbf{0.630} & 0.671 & 0.737 & 0.740 & 0.702 & 0.050 \\
\midrule
\texttt{IEKT}        & Neural & 0.743 & 0.711 & \textbf{0.681} & 0.718 & 0.714 & 0.713 & 0.022 \\
\bottomrule
\end{tabular}
}
\end{table}

Cross-model Pearson correlations of the fold-level AUC vectors confirm the synchrony (Table~\ref{tab:llm-kt-pearson}): all ten pairwise correlations among LLM models are $r \geq 0.944$ ($p < 0.02$), indicating that fold-level variance is driven predominantly by shared cohort composition rather than by model-specific architectural differences.

\begin{table}[!htbp]
\centering
\scriptsize
\setlength{\tabcolsep}{5pt}
\caption{Cross-model Pearson $r$ of fold-level AUC vectors for LLM-based KT models. All ten pairwise correlations are statistically significant ($p < 0.02$).}
\label{tab:llm-kt-pearson}
\begin{tabular}{@{}lccccc@{}}
\toprule
 & \texttt{Thinking-KT} & \texttt{LOKT} & \texttt{HISE-KT} & \texttt{CIKT} & \texttt{L-HAKT} \\
\midrule
\texttt{Thinking-KT} & 1.000 & 0.972 & 0.944 & 0.978 & 0.962 \\
\texttt{LOKT}        & 0.972 & 1.000 & 0.993 & 0.981 & 0.999 \\
\texttt{HISE-KT}     & 0.944 & 0.993 & 1.000 & 0.971 & 0.997 \\
\texttt{CIKT}        & 0.978 & 0.981 & 0.971 & 1.000 & 0.981 \\
\texttt{L-HAKT}      & 0.962 & 0.999 & 0.997 & 0.981 & 1.000 \\
\bottomrule
\end{tabular}
\end{table}

The fold-1 dip reflects a subject subpopulation assigned to fold~1 with shorter or lower-success interaction histories, precisely the condition under which the Wilson lower-bound prior contracts predictions toward 0.50 and the model loses discriminative power (Section~\ref{app:llm-kt-prior}). Per-sample probability logs from the faithful-run traces confirm this mechanism: fold~1 samples show greater prior contraction (mean blended $\hat{p}$ closer to 0.50) compared to other folds, consistent with the synchronised AUC drop across all five architectures. Cohort split details are in Appendix~\ref{app:cohort}.

\paragraph{Benchmarking implication.} The near-perfect inter-model AUC correlation shows that the five LLM-KT architectures are not informationally independent: they share a dominant failure mode that traces to the common backbone and prior mechanism rather than to prompt engineering. Reporting mean AUC without fold profiles would obscure this shared vulnerability.

\subsection{Test-Time Scaling Trace Analysis}
\label{app:llm-kt-tts}

\texttt{Thinking-KT} generates predictions at three reasoning-budget levels ($s \in \{1.0,\,1.5,\,2.0\}$, corresponding to 48, 72, and 96 reasoning tokens respectively) and aggregates by mean probability. Figure~\ref{fig:tts-trace} shows the annotated TTS trace for a representative sample from fold~0.

\begin{figure}[!htbp]
\centering
\scriptsize
\begin{minipage}{0.93\linewidth}
\begin{verbatim}
Sample: subject_d44103baa7 | Target: D111 (concept Rbh)
History (8 events): D103/Idm:C | D104/Apr:X | D105/Idm:X |
                    D106/Atb:C | D107/Que:C | D108/Atb:C |
                    D109/Wdr:X | D110/Nvr:X
History Prior: overall=0.500  recent=0.400  same_concept=0.333
-----------------------------------------------------------------
Scale x1.0  ->  confidence 0.43   (48 reasoning tokens,  6 steps)
Scale x1.5  ->  confidence 0.44   (72 reasoning tokens,  9 steps)
Scale x2.0  ->  confidence 0.45   (96 reasoning tokens, 12 steps)
-----------------------------------------------------------------
Aggregated probability: 0.44  |  Consistency: 3/3 (1.00)
Decision (threshold 0.72):    INCORRECT
Gold label:                   INCORRECT  [OK]
Prescription: "repeat the target concept with scaffolded review"
\end{verbatim}
\end{minipage}
\caption{Annotated Thinking-KT TTS trace (\texttt{buddybench\_nd03\_f0\_00000}). The confidence range across three budget levels is 0.02; the vote is unanimous. Per-sample latency: 23.5\,s.}
\label{fig:tts-trace}
\end{figure}

The confidence range across the three budget levels is 0.02 (0.43-0.45), and the TTS consistency rate is 1.0 (unanimous vote). Reasoning tokens increase $2\times$ from the minimum to maximum budget (48$\rightarrow$96 tokens, $6\rightarrow$12 trace steps), yet produce negligible change in the predicted probability. This is a direct consequence of the prior-dominated inference regime (Section~\ref{app:llm-kt-prior}): the history prior establishes a strong directional signal ($\pi \approx 0.40$, consistent with four recent errors and a same-concept success rate of 0.33), and the temperature-scaled LLM score provides only a minor additive correction. Additional compute budget at the current backbone scale cannot override the prior. This echoes \citet{snell2024tts}'s finding that test-time compute yields diminishing returns when contextual evidence is decisive, and \citet{wang2022consistency}'s observation that aggregating reasoning paths provides little gain when the model's response entropy is low.

\paragraph{Prescription output.} Beyond the binary prediction, \texttt{Thinking-KT} generates a structured clinical prescription (\emph{``repeat the target concept with scaffolded review''}) with a supporting rationale (\emph{``mixed or fragile recent evidence''}). These prescriptions are produced for every sample and represent an interpretability affordance entirely absent from neural KT baselines. Evaluating prescription quality against clinical outcomes is an open research direction enabled by BuddyBench's IRB-gated clinical endpoints.

\subsection{History Prior Decomposition}
\label{app:llm-kt-prior}

A five-component Bayesian history prior is applied uniformly across all five LLM-KT evaluations to regularise zero-shot predictions on the small ND-03 cohort, analogous to the history-prior calibration advocated by \citet{zhao2021calibrate} for few-shot-prompted classifiers. Because the prior is \emph{identical} for all five models, AUC comparisons among them reflect prompt-strategy differences rather than prior choice; the 0.012-point AUC range in Table~\ref{tab:llm-kt-metrics} is therefore interpretable as a controlled prompt-strategy comparison. These results should not be read as a direct reproduction of each model's out-of-box published performance on external benchmarks. Let $q$, $s$, $h$, $r$ denote the same-question, same-skill, overall history, and recent (last 3 events) response sequences respectively. Define the support saturation $\mathrm{supp}(c, k) = \min(1,\,|c|/k)$ with component thresholds $k_q{=}2$, $k_s{=}3$, $k_h{=}8$, $k_r{=}3$. The evidence strength, prior, and final calibrated score are:
\begin{align}
\epsilon &= 0.35\,\mathrm{supp}(q,2) + 0.25\,\mathrm{supp}(s,3) \nonumber \\
&\qquad  + 0.25\,\mathrm{supp}(h,8) + 0.15\,\mathrm{supp}(r,3) \label{eq:eps} \\
\pi &= \frac{\sum_{c}\,w_c\,\mathrm{WilsonLB}(c)}{\sum_c w_c}, \nonumber \\
&\qquad  w_c \in \{0.30,\,0.25,\,0.30,\,0.15\} \label{eq:prior} \\
\hat{p} &= (1-\alpha)\,\pi \;+\; \alpha\,\sigma\!\left(\tfrac{\mathrm{logit}(s_\mathrm{LLM})}{T}\right), \nonumber \\
&\qquad \alpha{=}0.45,\; T{=}4.0 \label{eq:blend}
\end{align}
where $\mathrm{WilsonLB}(c)$ is the lower bound of the Wilson score interval~\citep{wilson1927probable} and $\sigma$ is the sigmoid. The logit-temperature term ($T{=}4.0$) suppresses overconfident raw LLM label scores toward 0.50 before blending. \emph{Note:} when same-question evidence is absent ($|q|=0$), the $w_q{=}0.30$ weight is redistributed proportionally across the remaining three components; Equation~\eqref{eq:prior} shows the default weight configuration.

The key structural property is that $\mathrm{WilsonLB}$ is a \emph{conservative} estimator: for short sequences or mixed evidence, it assigns substantially lower probability than the sample mean. When $|h| < 8$, $\mathrm{supp}(h,8) < 1$ and all four component weights partially collapse, further contracting the prior toward 0.50. This explains the fold-1 synchronised dip (Section~\ref{app:llm-kt-folds}): if fold~1 contains a higher proportion of learners with short interaction histories, the prior provides weaker and more conservative signal for all LLM models simultaneously, regardless of their prompt-engineering differences.

\subsection{Prompt-Strategy Convergence Under Controlled Backbone}
\label{app:llm-kt-convergence}

To isolate the contribution of prompt engineering, we compare the five architecturally distinct strategies (clinical-weight prompting, option weighting, meta-path retrieval, hyperbolic teacher-student alignment, iterative profile refinement) under a controlled backbone (Qwen3-1.7B) and shared inference protocol. The AUC range across all five models is only 0.012 (0.700-0.712; Table~\ref{tab:llm-kt-metrics}). Pairwise paired $t$-tests across five folds find no significant difference for nine of the ten pairs; the single nominally significant pair (\texttt{LOKT} vs.\ \texttt{CIKT}, $p{=}0.046$) does not survive Bonferroni correction for ten comparisons ($\alpha_\mathrm{corr} = 0.005$). We therefore fail to reject the null hypothesis that the five prompt strategies are equivalent on BuddyBench.

This convergence result has two complementary interpretations. First, the shared Wilson lower-bound prior and backbone appear to impose a performance ceiling that prompt engineering at the 1.7B-parameter scale is unable to overcome on this benchmark and cohort, consistent with saturation effects observed in chain-of-thought complexity scaling~\citep{wang2022consistency}, though a larger backbone or cohort would be needed to confirm the generality of this ceiling. Second, BuddyBench Task~1's 153-drill, 28-concept space offers limited semantic variety for meta-path retrieval (\texttt{HISE-KT}) or option-sequence encoding (\texttt{LOKT}) to outperform simpler strategies. Both effects likely contribute.

\paragraph{Caveat.} With $n{=}5$ folds, paired $t$-tests are underpowered for detecting small effects (Cohen's $d < 0.5$). The null result should be read as ``no evidence of difference'' rather than ``evidence of equivalence.'' Cross-site replication with a larger participant pool would be needed to establish practical equivalence formally.

\subsection{Inference Efficiency Profile}
\label{app:llm-kt-efficiency}

All five LLM-KT models complete a full fold evaluation in 45.7-49.9\,s (Table~\ref{tab:llm-kt-metrics}), representing \emph{inference-only} runtime with no gradient-based training. The per-sample latency for \texttt{Thinking-KT} in TTS mode is approximately 23.5\,s (from the faithful-run record), driven by three sequential LLM calls at different reasoning budgets. Non-TTS models (\texttt{LOKT}, \texttt{HISE-KT}, \texttt{CIKT}, \texttt{L-HAKT}) require a single LLM call per sample and achieve lower wall times (45.7-49.0\,s/fold). For comparison, neural KT models in this benchmark require 2-8\,h of GPU training per fold before inference (Appendix~\ref{app:computation}).

The zero-shot nature of LLM-KT inference is directly relevant to BuddyBench's privacy-constrained setting: LLM models can be evaluated on ND-03 data without any parameter updates, avoiding the risk of overfitting to sensitive training data inherent in any gradient-based adaptation. This makes LLM-KT models appealing for settings where retraining on sensitive paediatric data is restricted by IRB constraints or impractical given small cohort sizes.

\section{Task~1: Tokenized Masked-Sequence Protocol}
\label{app:kt-task1}

Task~1 knowledge tracing uses a tokenized masked-sequence representation rather than a separable tabular feature set. Each participant's drill history is encoded as an ordered token sequence over the 153-item drill bank; the target drill is masked at its sequence position and the model predicts its correctness from the surrounding token context.

The \texttt{F3\_tokenized\_masked} identifier used throughout this paper refers to this protocol under the F3 feature augmentation level. At F3, per-drill summary statistics, prior accuracy, attempt count, and skill-coverage fraction, are encoded as discrete token extensions appended to each drill token. Because these statistics are part of the token vocabulary rather than separable tabular columns, lower tiers (F0--F2) correspond to progressively sparser token contexts and are not amenable to the column-additive ablation readout used for Tasks~2--4. Task~1 is therefore evaluated at F3 throughout the paper.

Concretely, each training sample consists of a participant's interaction sequence with the target response masked; the model is trained to recover the masked correctness label from the preceding history. Evaluation follows a participant-disjoint 5-fold subject-split (seed~42): models are trained on the pooled sequences of training-partition participants and evaluated on held-out participants' masked sequences. This design prevents within-participant memorization and ensures that leaderboard AUC reflects generalization to unseen learners rather than per-student history recall.

\section{Feature Ablation: F0-F3}
\label{app:feature-ablation}

BuddyBench defines a shared feature-ablation hierarchy: F0 uses only drill and learner IDs; F1 adds demographic and baseline clinical variables; F2 adds task-legal BuddyPlan self-report factors; F3 adds drill-summary statistics (prior accuracy, attempt count, skill coverage). An analogous scheme applies to T3--T4, where temporal leakage constraints exclude post-intervention variables throughout. Table~\ref{tab:feature-ablation} consolidates the feature-tier ablation for Tasks~2--4, where each feature group is independently additive over a shared tabular feature space. Task~1 uses a tokenized masked-sequence architecture (Appendix~\ref{app:kt-task1}) in which F3 encodes drill-summary statistics as part of the input token vocabulary rather than as separable tabular columns, and is therefore evaluated at F3 throughout. Task~3 uses F2 as the primary reported setting, while F3 is retained only as an auxiliary ablation point.

\paragraph{Feature Group Definitions per Task.}
Table~\ref{tab:feature-spec} specifies which content each F-tier adds and whether it is
task-legal for each task. F3 content differs by task: drill-summary statistics are
behavioral features task-legal for T1-T2 and are also included for T3-T4 as leakage-safe
pre-intervention summaries. BuddyPlan items exist in two variants: the 52-item ND-03 form
used in T1--T2, and the 18-item ND-02 subset encoded under the V2\_PLAN prefix for T3--T4.
Post-treatment variables (V6\_\*, Diff\_\*) are excluded from all tiers for Tasks~3--4.

\begin{table}[!htbp]
\centering
\scriptsize
\setlength{\tabcolsep}{3pt}
\renewcommand{\arraystretch}{0.95}
\caption{Feature-set content by task. $\checkmark$ = task-legal, $\circ$ = partial, - = not applicable.}
\label{tab:feature-spec}
\begin{tabular*}{\columnwidth}{@{\extracolsep{\fill}}p{0.09\columnwidth}p{0.50\columnwidth}cccc@{}}
\toprule
\textbf{Tier} & \textbf{Content added} & \textbf{T1} & \textbf{T2} & \textbf{T3} & \textbf{T4} \\
\midrule
F0 & Drill ID + learner ID only & $\checkmark$ & $\checkmark$ & - & - \\
F1 & \shortstack[l]{+ demographics +\\baseline SRS-2/VABS-II} & $\checkmark$ & $\checkmark$ & $\checkmark$ & $\checkmark$ \\
F2 & \shortstack[l]{+ BuddyPlan self-report\\(52 / 18 items)} & $\checkmark$ & $\checkmark$ & $\circ$ & $\circ$ \\
F3 & \shortstack[l]{+ drill summaries\\(acc., attempts, cov.)} & $\checkmark$ & $\checkmark$ & $\checkmark$ & $\checkmark$ \\
\bottomrule
\end{tabular*}
\end{table}

\begin{table*}[!htbp]
\centering
\scriptsize
\setlength{\tabcolsep}{2pt}
\renewcommand{\arraystretch}{0.95}
\caption{Feature-tier ablation for Tasks~2--4 (best-model primary metric per tier). Task~1 uses a tokenized masked-sequence architecture evaluated at F3 throughout; see Appendix~\ref{app:kt-task1}.}
\label{tab:feature-ablation}
\begin{tabular*}{\textwidth}{@{\extracolsep{\fill}}p{0.07\textwidth}p{0.22\textwidth}p{0.18\textwidth}p{0.15\textwidth}p{0.27\textwidth}@{}}
\toprule
\textbf{Set} & \textbf{Content added} & \textbf{T2 RecSys} & \textbf{T3 Clinical} & \textbf{T4 Causal} \\
\midrule
F0 & IDs only
& \shortstack[l]{\texttt{DiffDiv}\\0.322 / 0.431}
& -
& - \\
F1 & + demographics + baseline clinical
& \shortstack[l]{\texttt{DiffDiv}\\0.322 / 0.431}
& \shortstack[l]{\texttt{tabm}\\0.859}
& \shortstack[l]{\texttt{TARNet}\\0.191 / 126.944 / 6.553} \\
F2 & + BuddyPlan
& \shortstack[l]{\texttt{DiffDiv}\\0.322 / 0.431}
& \shortstack[l]{\texttt{tabm}\\0.860}
& \shortstack[l]{\texttt{CausalForest} / \texttt{RLearner} / \texttt{ATEOnly}\\0.205 / 172.309 / 8.449} \\
F3 & + drill summaries
& \shortstack[l]{\texttt{DiffDiv}\\0.322 / 0.431}
& \shortstack[l]{\texttt{elastic\_net}\\0.853}
& \shortstack[l]{\texttt{CausalForest} / \texttt{ATEOnly} / \texttt{ATEOnly}\\0.240 / 181.368 / 9.397} \\
\bottomrule
\end{tabular*}
\end{table*}

\section{BuddyBench-Sim Benchmark Results}
\label{app:synthetic-benchmark}

Tables~\ref{tab:synthetic-t1}-\ref{tab:synthetic-t4} summarize BuddyBench-Sim evidence for Tasks~1--4. Table~\ref{tab:synthetic-t1} reports the current Task~1 synthetic leaderboard under the recommended participant-split masked F3 protocol; several rows remain near chance, which reflects transfer difficulty rather than a data quality issue. For T2, ``Retention'' is the ratio of synthetic to real primary metric (R@10). T4 reports endpoint-level residualized R-loss and PEHE rather than a retention ratio because metric scales differ between real and synthetic outcomes. For T3, the synthetic results are shown as auxiliary F3 utility diagnostics rather than a matched counterpart to the real-data F2 protocol, and for T4, PEHE is available only for the synthetic data-generating process.

\paragraph{ML utility score definition.}
The \emph{aggregate ML utility score} of 0.765 reported in the main paper (\S\ref{sec:synthetic-readiness}) is the mean of task-level utility measures across all tasks where cross-cohort comparison is computable: (1) Spearman rank correlation of model AUC rankings between BuddyBench-Sim and real BuddyBench for T1 ($\rho = 0.746$, 15 models); (2--3) Spearman rank correlation of model recommendation rankings for T2 on R@10 ($\rho = 0.675$) and NDCG@10 ($\rho = 0.752$, both 19 models); and (4) the normalized T3 best-model retention ratio $\min(1,\, \text{real AUPRC} / \text{sim AUPRC})$ ($= 0.860/0.979 \approx 0.879$). Tasks~1 and 2 each contribute equally, giving $\bar{\rho} = (0.746 + 0.714 + 0.879) / 3$, weighted by task. The full per-task breakdown appears in Tables~\ref{tab:rank-preservation} and~\ref{tab:sim-gap-summary}. T4 is excluded from this aggregate because real and synthetic residualized R-loss values differ by orders of magnitude due to outcome scale misalignment.

\begin{table}[!htbp]
\centering
\scriptsize
\setlength{\tabcolsep}{4pt}
\renewcommand{\arraystretch}{0.95}
\caption{Task~1 BuddyBench-Sim leaderboard under the recommended participant-split masked \texttt{F3\_tokenized\_masked} protocol (see Appendix~\ref{app:kt-task1}). Mean $\pm$ SD across 5 folds (seed 42).}
\label{tab:synthetic-t1}
\begin{tabular*}{\columnwidth}{@{\extracolsep{\fill}}lccc@{}}
\toprule
\textbf{Model} & \textbf{ACC} & \textbf{AUC} & \textbf{F1} \\
\midrule
\texttt{AKT} & $0.54 \pm 0.03$ & $0.55 \pm 0.03$ & $0.60 \pm 0.05$ \\
\texttt{DKVMN} & $0.51 \pm 0.01$ & $0.51 \pm 0.01$ & $0.48 \pm 0.06$ \\
\texttt{Deep-IRT} & $0.50 \pm 0.01$ & $0.50 \pm 0.01$ & $0.50 \pm 0.07$ \\
\texttt{DKT+} & $0.57 \pm 0.00$ & $0.59 \pm 0.01$ & $0.57 \pm 0.01$ \\
\texttt{DKT} & $0.57 \pm 0.00$ & $0.59 \pm 0.01$ & $0.57 \pm 0.01$ \\
\texttt{DTransformer} & $0.50 \pm 0.01$ & $0.50 \pm 0.01$ & $0.34 \pm 0.19$ \\
\texttt{ExTRAKT} & $0.50 \pm 0.01$ & $0.50 \pm 0.01$ & $0.29 \pm 0.27$ \\
\texttt{FOLIBIKT} & $0.57 \pm 0.00$ & $0.59 \pm 0.01$ & $0.57 \pm 0.01$ \\
\texttt{GKT} & $0.58 \pm 0.00$ & $0.58 \pm 0.01$ & $0.57 \pm 0.01$ \\
\texttt{Hawkes} & $0.51 \pm 0.01$ & $0.51 \pm 0.00$ & $0.48 \pm 0.04$ \\
\texttt{HCGKT} & $0.58 \pm 0.00$ & $0.61 \pm 0.00$ & $0.57 \pm 0.01$ \\
\texttt{HISE-KT} & $0.49 \pm 0.06$ & $0.55 \pm 0.06$ & $0.35 \pm 0.08$ \\
\texttt{IEKT} & $0.58 \pm 0.00$ & $0.62 \pm 0.00$ & $0.57 \pm 0.02$ \\
\texttt{KnowNet} & $0.58 \pm 0.01$ & $0.64 \pm 0.01$ & $0.58 \pm 0.03$ \\
\texttt{KQN} & $0.57 \pm 0.01$ & $0.57 \pm 0.00$ & $0.57 \pm 0.02$ \\
\texttt{L-HAKT} & $0.49 \pm 0.07$ & $0.54 \pm 0.06$ & $0.27 \pm 0.07$ \\
\texttt{LEFOKT-AKT} & $0.50 \pm 0.01$ & $0.51 \pm 0.02$ & $0.53 \pm 0.10$ \\
\texttt{LKT} & $0.62 \pm 0.01$ & $0.55 \pm 0.00$ & $0.00 \pm 0.00$ \\
\texttt{LOKT} & $0.49 \pm 0.06$ & $0.54 \pm 0.06$ & $0.30 \pm 0.08$ \\
\texttt{AKT-Custom} & $0.56 \pm 0.03$ & $0.56 \pm 0.03$ & $0.48 \pm 0.16$ \\
\texttt{ATDKT} & $0.57 \pm 0.01$ & $0.59 \pm 0.01$ & $0.56 \pm 0.01$ \\
\texttt{ATKT} & $0.58 \pm 0.00$ & $0.59 \pm 0.00$ & $0.57 \pm 0.01$ \\
\texttt{ATKTFix} & $0.52 \pm 0.01$ & $0.53 \pm 0.01$ & $0.55 \pm 0.03$ \\
\texttt{CIKT} & $0.47 \pm 0.07$ & $0.54 \pm 0.05$ & $0.21 \pm 0.06$ \\
\texttt{CSKT} & $0.56 \pm 0.03$ & $0.56 \pm 0.03$ & $0.46 \pm 0.23$ \\
\texttt{PEBG} & $0.58 \pm 0.01$ & $0.63 \pm 0.01$ & $0.57 \pm 0.01$ \\
\texttt{QDKT} & $0.58 \pm 0.01$ & $0.64 \pm 0.01$ & $0.57 \pm 0.01$ \\
\texttt{QIKT} & $0.58 \pm 0.00$ & $0.62 \pm 0.03$ & $0.57 \pm 0.00$ \\
\texttt{ReKT} & $0.51 \pm 0.01$ & $0.51 \pm 0.01$ & $0.31 \pm 0.25$ \\
\texttt{RKT} & $0.50 \pm 0.01$ & $0.51 \pm 0.01$ & $0.18 \pm 0.26$ \\
\texttt{RobustKT} & $0.51 \pm 0.03$ & $0.52 \pm 0.04$ & $0.26 \pm 0.20$ \\
\texttt{SAINT} & $0.57 \pm 0.01$ & $0.61 \pm 0.01$ & $0.58 \pm 0.01$ \\
\texttt{SAINT+} & $0.57 \pm 0.00$ & $0.61 \pm 0.01$ & $0.57 \pm 0.00$ \\
\texttt{SAKT} & $0.50 \pm 0.01$ & $0.50 \pm 0.01$ & $0.41 \pm 0.27$ \\
\texttt{SimpleKT} & $0.50 \pm 0.01$ & $0.51 \pm 0.01$ & $0.35 \pm 0.26$ \\
\texttt{SKVMN} & $0.50 \pm 0.01$ & $0.51 \pm 0.00$ & $0.43 \pm 0.23$ \\
\texttt{SparseKT} & $0.50 \pm 0.01$ & $0.50 \pm 0.01$ & $0.19 \pm 0.24$ \\
\texttt{StableKT} & $0.57 \pm 0.00$ & $0.60 \pm 0.01$ & $0.56 \pm 0.01$ \\
\texttt{Thinking-KT} & $0.46 \pm 0.06$ & $0.54 \pm 0.06$ & $0.13 \pm 0.05$ \\
\texttt{UKT} & $0.58 \pm 0.00$ & $0.58 \pm 0.01$ & $0.57 \pm 0.01$ \\
\bottomrule
\end{tabular*}
\end{table}

\begin{table}[!htbp]
\centering
\scriptsize
\setlength{\tabcolsep}{2pt}
\renewcommand{\arraystretch}{0.92}
\caption{Task~2 BuddyBench-Sim vs.\ real ND-03 under warm-start F3, part 1. Retention uses R@10.}
\label{tab:synthetic-t2}
\resizebox{\columnwidth}{!}{%
\begin{tabular}{lcccccc}
\toprule
\textbf{Model} & \textbf{Real} & \textbf{S@5} & \textbf{S@10} & \textbf{S@20} & \textbf{SN@10} & \textbf{Ret.} \\
\midrule
\texttt{CSRec} & 0.3015 & \textbf{0.2110} & \textbf{0.3565} & \textbf{0.5784} & \textbf{0.5048} & 1.182 \\
\texttt{C3Rec} & 0.2648 & 0.1772 & 0.3130 & 0.4988 & 0.4402 & 1.182 \\
\texttt{NILUS} & 0.2624 & 0.1823 & 0.3031 & 0.4782 & 0.4708 & 1.155 \\
\texttt{DiffDiv} & 0.3221 & 0.1426 & 0.2671 & 0.4716 & 0.3919 & 0.829 \\
\texttt{PreferDiff} & 0.3143 & 0.1453 & 0.2696 & 0.4716 & 0.3950 & 0.858 \\
\texttt{SASRec} & 0.2620 & 0.1181 & 0.2691 & 0.4543 & 0.3435 & 1.027 \\
\texttt{SIGMA} & 0.2594 & 0.0965 & 0.2357 & 0.4301 & 0.3060 & 0.909 \\
\texttt{RecVAE} & 0.2027 & 0.1478 & 0.2732 & 0.4708 & 0.3921 & 1.348 \\
\texttt{FailureRate} & 0.3168 & 0.1453 & 0.2694 & 0.4716 & 0.3972 & 0.850 \\
\texttt{TS-Rec} & 0.3142 & 0.1410 & 0.2696 & 0.4716 & 0.3723 & 0.858 \\
\bottomrule
\end{tabular}}
\end{table}

\begin{table}[!htbp]
\centering
\scriptsize
\setlength{\tabcolsep}{2pt}
\renewcommand{\arraystretch}{0.92}
\caption{Task~2 BuddyBench-Sim vs.\ real ND-03 under warm-start F3, part 2.}
\label{tab:synthetic-t2-continued}
\resizebox{\columnwidth}{!}{%
\begin{tabular}{lcccccc}
\toprule
\textbf{Model} & \textbf{Real} & \textbf{S@5} & \textbf{S@10} & \textbf{S@20} & \textbf{SN@10} & \textbf{Ret.} \\
\midrule
\texttt{Cold-Start} & 0.2298 & 0.1342 & 0.2357 & 0.3866 & 0.3304 & 1.026 \\
\texttt{LightGCN} & 0.1822 & 0.0768 & 0.1672 & 0.3166 & 0.2082 & 0.918 \\
\texttt{SimGCL} & 0.1546 & 0.0890 & 0.1646 & 0.3255 & 0.2398 & 1.065 \\
\texttt{Disco} & 0.2714 & 0.0809 & 0.1568 & 0.3093 & 0.2314 & 0.578 \\
\texttt{BERT4Rec} & 0.1689 & 0.0646 & 0.1055 & 0.2968 & 0.2134 & 0.625 \\
\texttt{GRACE} & 0.1028 & 0.0468 & 0.0761 & 0.1713 & 0.1297 & 0.740 \\
\bottomrule
\end{tabular}}
\end{table}

\begin{table}[!htbp]
\centering
\scriptsize
\setlength{\tabcolsep}{4pt}
\renewcommand{\arraystretch}{0.95}
\caption{Task~3 BuddyBench-Sim auxiliary utility comparison vs.\ real ND-02, using the best dependency-available full-size F3 row (single run; the 3-seed averaged value at $n{=}1{,}000$ in Figure~\ref{fig:scaling-combined} is 0.981).}
\label{tab:synthetic-t3}
\begin{tabular*}{\columnwidth}{@{\extracolsep{\fill}}p{0.28\columnwidth}ccc@{}}
\toprule
\textbf{Target} & \textbf{Real best AUPRC} & \textbf{Sim best AUPRC} & \textbf{Retention} \\
\midrule
Non-response risk & 0.860 & 0.979 & 1.139 \\
\bottomrule
\end{tabular*}
\end{table}

\begin{table}[!htbp]
\centering
\scriptsize
\setlength{\tabcolsep}{3pt}
\renewcommand{\arraystretch}{0.94}
\caption{Task~4 BuddyBench-Sim vs.\ real ND-02.}
\label{tab:synthetic-t4}
\resizebox{\columnwidth}{!}{%
\begin{tabular}{lcccc}
\toprule
\textbf{Endpoint} & \textbf{Real R-loss} & \textbf{Sim R-loss} & \textbf{PEHE} & \textbf{Best} \\
\midrule
CGI-S/I & 0.240 & 54.808 & 5.768 & \texttt{BART} \\
SRS-2 & 181.368 & 1092.882 & 21.934 & \texttt{GRFCausalForest} \\
VABS  & 9.397 & 658.939 & 17.592 & \texttt{CausalPFN} \\
\bottomrule
\end{tabular}}
\end{table}

\subsection{BuddyBench-Sim Rank Preservation}
\label{app:synthetic-rank}

Table~\ref{tab:rank-preservation} reports the Spearman rank correlation between real and synthetic model rankings for tasks where multiple models were evaluated on both cohorts under comparable protocols. Task~2 achieves moderate-to-strong rank preservation ($\rho = 0.675$, $p = 0.002$ on R@10; $\rho = 0.752$, $p < 0.001$ on NDCG@10 across 19 models), supporting the use of BuddyBench-Sim for relative method screening before applying for real-data access. Task~1 AUC rank correlation is moderate ($\rho = 0.746$, $p = 0.001$ across 15 models, five-fold averages), indicating meaningful but imperfect rank preservation. Tasks~3--4 are not amenable to model-rank correlation: Task~3 has a single best-model retention point (AUPRC retention $= 1.14$), and Task~4 real and synthetic R-loss values differ by two orders of magnitude owing to outcome scale differences. The Task~2 rank correlation result implies that near-chance synthetic Task~1 rows should not be read as predicting near-chance real-data performance; the transfer-difficulty signal in Table~\ref{tab:synthetic-t1} reflects uneven schema compatibility rather than a general claim about real-cohort model ordering.

\begin{table}[!htbp]
\centering
\scriptsize
\setlength{\tabcolsep}{4pt}
\renewcommand{\arraystretch}{0.95}
\caption{BuddyBench-Sim rank preservation: Spearman $\rho$ between real and synthetic model rankings. Dashes indicate tasks where rank correlation is not computable; see text.}
\label{tab:rank-preservation}
\begin{tabular*}{\columnwidth}{@{\extracolsep{\fill}}lcccc@{}}
\toprule
\textbf{Task} & \textbf{Metric} & \textbf{$n$ models} & \textbf{Spearman $\rho$} & \textbf{$p$-value} \\
\midrule
T1 KT       & AUC     & 15 & $0.746$ & $0.001$ \\
T2 RecSys   & R@10    & 19 & $0.675$ & $0.002$ \\
T2 RecSys   & NDCG@10 & 19 & $0.752$ & $<$0.001 \\
T3 Clinical & AUPRC   & 1  & -      & - \\
T4 Causal   & R-loss  & - & -      & - \\
\bottomrule
\end{tabular*}
\end{table}

\subsection{Real-to-Synthetic Performance Gap: Task-Directional Asymmetry}
\label{app:synthetic-gap}

Performance differences between BuddyBench and BuddyBench-Sim are \emph{task-directional} rather than uniformly favorable or unfavorable. We refer to these differences collectively as the \emph{synthetic gap}: they reflect structural mismatches between each task's real-cohort properties and the PI-VAE synthetic generation process, not an undifferentiated global fidelity failure. Table~\ref{tab:sim-gap-summary} summarizes the direction and primary cause per task using the same model across both cohorts wherever possible.

\begin{table}[!htbp]
\centering
\scriptsize
\setlength{\tabcolsep}{3pt}
\renewcommand{\arraystretch}{0.95}
\caption{Synthetic gap summary: direction and primary structural cause by task. Metric values use the real-best model evaluated on both cohorts (same model for T1-T3; T4 scales are incomparable).}
\label{tab:sim-gap-summary}
\begin{tabular*}{\columnwidth}{@{\extracolsep{\fill}}p{0.12\columnwidth}p{0.13\columnwidth}p{0.08\columnwidth}p{0.10\columnwidth}p{0.10\columnwidth}p{0.34\columnwidth}@{}}
\toprule
\textbf{Task} & \textbf{Metric} & \textbf{Real} & \textbf{Sim} & \textbf{Dir.} & \textbf{Primary cause} \\
\midrule
T1 KT       & AUC (\texttt{PEBG})   & 0.72 & 0.63 & Sim$\downarrow$ & Near-complete real matrix (88.1\% fill) not preserved in synthesis \\
T2 RecSys   & R@10 (\texttt{DiffDiv}) & 0.322 & 0.267 & Mixed & CF methods sim-inflated; domain heuristics sim-deflated \\
T3 Clinical & AUPRC (\texttt{tabm})  & 0.860 & 0.979 & Sim$\uparrow$  & Synthetic regime ${\approx}12{\times}$ larger ($n{=}1{,}000$ vs.\ 83) with smoother distribution \\
T4 Causal   & R-loss                 & 0.240 & 54.8  & N/A & Outcome scales incomparable across real and synthetic cohorts \\
\bottomrule
\end{tabular*}
\end{table}

\begin{figure*}[!htbp]
\centering
\includegraphics[width=\linewidth]{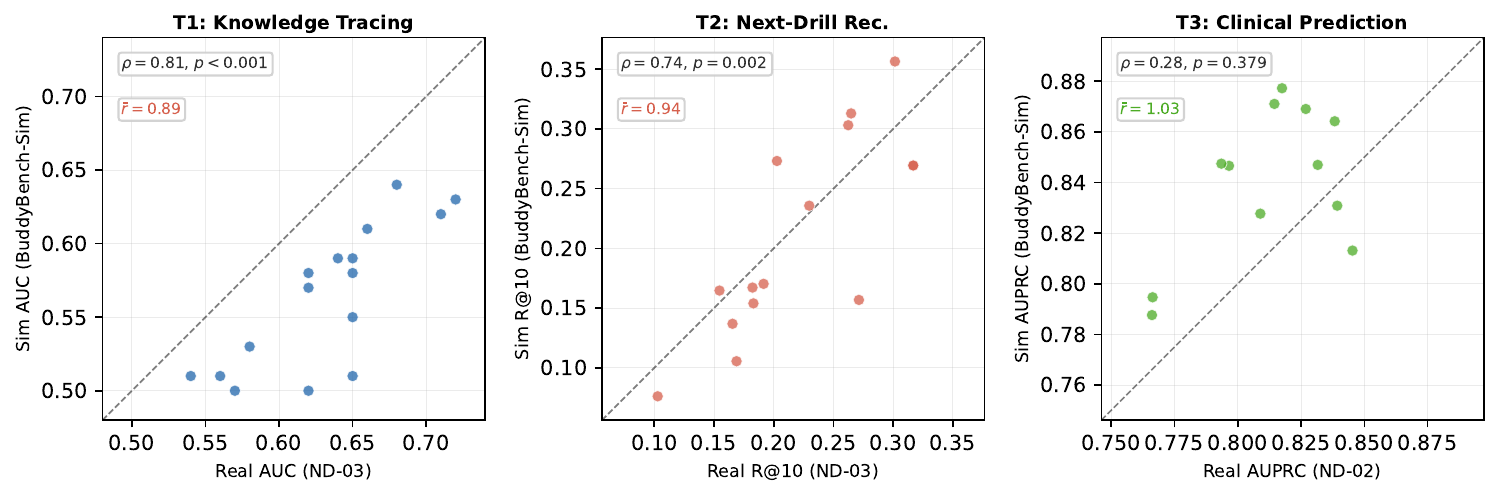}
\caption{Real vs.\ BuddyBench-Sim performance scatter for T1 (AUC), T2 (R@10), and T3 (AUPRC).
Each point is one model evaluated on both cohorts under comparable protocols.
The dashed diagonal is the identity line ($y{=}x$); points above it indicate synthetic inflation and below indicate synthetic degradation.
$\bar{r}$: mean Sim/Real performance ratio. $\rho$: Spearman rank correlation ($p{=}0.001$ for T1, $p{<}0.001$ for T2).
T4 is omitted: residualized R-loss scales differ by orders of magnitude across real and synthetic outcomes (Table~\ref{tab:sim-gap-summary}).}
\label{fig:real-sim-heatmap}
\end{figure*}

\paragraph{T1 (Knowledge Tracing): sim-degraded.}
Real ND-03 has an 88.1\% drill-response fill rate and 137.4 average attempts per participant on a 153-item bank, a near-complete interaction matrix that provides rich within-participant history for KT models. BuddyBench-Sim does not replicate this density: its synthetic fill structure is smoother but sparser, placing KT models in a setting where temporal sequence signals are weaker. Consequently, models that achieve AUC 0.62-0.72 on real ND-03 commonly fall to 0.50-0.64 on BuddyBench-Sim, with several rows near chance (AUC $\approx 0.50$; Table~\ref{tab:synthetic-t1}). The moderate rank-preservation result ($\rho = 0.746$; Table~\ref{tab:rank-preservation}) confirms that relative ordering is meaningful despite the absolute gap: BuddyBench-Sim T1 supports valid method screening but should not be used as a difficulty proxy for real-data KT.

\paragraph{T2 (Next-Drill Recommendation): model-family-dependent.}
T2 shows a model-family-dependent gap rather than a uniform direction. Collaborative filtering and neural methods (\texttt{BERT4Rec}, \texttt{RecVAE}, \texttt{LightGCN}, \texttt{SimGCL}) score substantially higher on BuddyBench-Sim (e.g.\ \texttt{BERT4Rec} NDCG@10: real $0.25$ vs.\ sim $0.40$; \texttt{RecVAE} R@10: real $0.20$ vs.\ sim $0.27$), while domain-aware heuristics score lower (\texttt{DiffDiv} R@10: real $0.322$ vs.\ sim $0.267$; \texttt{FailureRate} NDCG@10: real $0.43$ vs.\ sim $0.40$). This divergence reflects the synthetic cohort's more generic interaction structure, which is favorable to CF-style collaborative signal but less representative of the practice-to-criterion patterns that domain-aware heuristics are designed to exploit. Rank preservation is moderate-to-strong ($\rho = 0.675$--$0.752$; Table~\ref{tab:rank-preservation}), confirming that BuddyBench-Sim T2 supports valid method screening despite this family-level divergence.

\paragraph{T3 (Clinical Prediction): sim-inflated.}
The gap is large and directional: best real AUPRC is 0.860 (\texttt{tabm}) versus best synthetic AUPRC of 0.979 (Table~\ref{tab:t3-response-synthetic-full}, single run; three-seed average 0.981 in Figure~\ref{fig:scaling-combined}). Two structural factors drive this inflation: (a) the synthetic cohort is more than twelve times the real FAS ($n{=}83$), collapsing fold-level variance and enabling tighter model fits; and (b) the PI-VAE synthetic distribution is smoother than the real clinical distribution, reducing label noise. This inflation is expected and interpretable as evidence of the benchmark's scalable development role: the scaling analysis (Appendix~\ref{app:scaling}) shows T3 AUPRC growing monotonically from 0.887 at $n{=}160$ to 0.981 at $n{=}1{,}000$ on BuddyBench-Sim, demonstrating that the clinical-prediction task is solvable with sufficient data, a result unattainable with real data alone. The synthetic inflation is therefore not a fidelity failure but a confirmation that data quantity, not model capacity, limits real-data T3 performance.

\paragraph{T4 (Causal Inference): not directly comparable.}
Real and synthetic residualized R-loss values differ by two to three orders of magnitude (e.g.\ CGI-S/I: real $0.240$ vs.\ sim $54.8$; SRS-2: real $181$ vs.\ sim $1{,}093$) because real and synthetic outcome scales are not aligned. Cross-cohort R-loss comparison is therefore not interpretable as a performance gap. Table~\ref{tab:synthetic-t4} provides PEHE as a synthetic-specific metric uniquely computable from the data-generating process; it is reported as a supplementary diagnostic rather than as a retention comparison against the real leaderboard.

\paragraph{Summary and practical guidance.}
The synthetic gap is task-specific and structurally interpretable: sim-degraded on T1 (dense-matrix challenge not preserved), model-family-dependent on T2 (CF vs.\ domain-heuristic divergence), sim-inflated on T3 (large-$n$ smoother regime), and not comparable on T4 (scale mismatch). In all tasks, BuddyBench-Sim functions as a valid executability screen and method-ranking diagnostic, but synthetic performance scores should not be interpreted as estimates of real-cohort difficulty. The recommended use pattern is: screen and rank methods using BuddyBench-Sim, then validate finalists under the dual-IRB real-data access protocol.

\section{Benchmark Limitations}
\label{app:limitations}

\paragraph{Single-site, small cohorts.}
Both cohorts come from a single institution. ND-03 has a 2.4:1 M:F ratio and ND-02 a 13:1 ratio, both characteristic of ASD/SCD clinical trial populations but not representative of the general pediatric population. The ND-03 diagnostic distribution spans six categories (SCD 29.5\%, ASD 27.9\%, high-risk 27.3\%, ADHD 13.1\%, ID 1.1\%, other/missing 1.1\%), whereas ND-02 is concentrated in ASD/SCD. Between-diagnosis comparisons and any subgroup claims should therefore be read conservatively.

\paragraph{Sparse drill coverage in ND-02.}
The 32.0\% drill fill rate in ND-02 (vs.\ 88.1\% in ND-03) reflects the narrower RCT protocol and typical compliance heterogeneity in a 12-week pediatric intervention. Task~1 is therefore anchored to ND-03, where interaction histories are denser and behavioral coverage is broader; ND-02 is not part of the primary KT leaderboard in this paper.

\paragraph{RCT underpowering for heterogeneous treatment-effect detection.}
The $n = 86$ ITT sample provides reasonable power to detect average treatment effects but is severely underpowered for heterogeneity estimation. The narrow margin between \texttt{CausalForest} and constant-effect baselines in Table~\ref{tab:t4-causal-extended} ($\Delta = 0.009$ CGI-S/I residualized R-loss) is consistent with this limitation. Task~4 benchmarks causal-inference \emph{methodology} in small clinical RCTs rather than providing evidence that meaningful heterogeneity exists in the target population.

\paragraph{Offline recommendation proxy.}
Task~2 uses an offline top-$k$ ranking proxy. Drill ordering is inferred from the column order of the wide data matrix rather than from timestamps, which are withheld to reduce re-identification risk. Results therefore reflect an offline ranking protocol rather than a true online adaptive therapeutic policy evaluation.

\paragraph{Synthetic benchmark coverage.}
BuddyBench-Sim task-level diagnostics are reported for T2 and T4, with T3 retained as an auxiliary F3 utility check in Tables~\ref{tab:synthetic-t2}-\ref{tab:synthetic-t4}. Table~\ref{tab:synthetic-t1} shows that synthetic KT utility is uneven across architectures, with several rows remaining near chance. The main-paper aggregate statistics (correlation preservation $\geq 0.968$, membership-inference accuracy $\leq 0.52$) characterize \emph{statistical} fidelity; task-level prediction utility is a separate question tracked in these tables.

\paragraph{Feature ablation coverage.}
T2 ablation results show that the best-model R@10 is flat across F0-F3, confirming that top-ranked difficulty-aware heuristics are insensitive to learner-profile feature richness. T3 ablation results show that AUPRC peaks at F2 (0.860) rather than F3 (0.853), suggesting that additional drill-summary features add noise rather than signal at $n = 83$. T4 ablation results show increasing residualized R-loss with richer feature tiers on all three endpoints, consistent with the underpowered small-RCT regime.

\section{Benchmark Difficulty Calibration}
\label{app:difficulty}

A credible benchmark should neither be trivially solved nor arbitrarily hard. This section documents evidence that Tasks~1--4 remain substantively unsaturated by current models.

\paragraph{Task~1.}
The highest five-fold AUC across the 36-model real-data leaderboard is 0.72 (\texttt{PEBG}), well below ceiling. AUC values span 0.54-0.72 (a 0.18-point range), indicating strong discriminability across model families. Fold-wise standard deviations reach 0.12 for individual models (e.g.\ \texttt{LOKT}), reflecting genuine variability in knowledge-tracing difficulty rather than measurement noise.

\paragraph{Task~2.}
The best real-cohort R@10 under the warm-start F3 protocol is 0.322, meaning the top-ranked model retrieves roughly one-third of all relevant next drills within 10 recommendations. The best NDCG@10 (0.431) likewise leaves substantial headroom. Heuristic methods remain competitive with learned recommenders, suggesting that learner-profile signal is not yet being fully exploited.

\paragraph{Task~3.}
The best AUPRC (0.860) is a $+$0.113 improvement over the prevalence baseline (0.747), indicating real but modest pre-treatment risk-ranking signal. The best AUROC is 0.614, showing that threshold-dependent binary classification is substantially harder than ranking. The AUPRC-AUROC gap itself signals task difficulty: models must rank high-risk individuals correctly despite majority-class imbalance, and the small $n = 83$ limits generalization.

\paragraph{Task~4.}
The best CGI-S/I residualized R-loss is 0.240 (\texttt{CausalForest}) against the constant-effect baseline \texttt{ATEOnly} at 0.249, a margin of $\Delta = 0.009$ across five outer folds. This narrow gap is consistent with the fundamental challenge of heterogeneous treatment-effect estimation at $n = 86$ in a short pediatric RCT, where individualized treatment differences are small relative to outcome noise. Task~4 benchmarks causal-inference methodology rather than easy discrimination.

\section{Cohort Characteristics and Split Construction}
\label{app:cohort}

\subsection{Cohort Summary}

Table~\ref{tab:cohort-summary} summarizes the key characteristics of the two real cohorts. Both come from a single institution and are not representative of the general pediatric population.

\begin{table}[!htbp]
\centering
\scriptsize
\setlength{\tabcolsep}{3pt}
\renewcommand{\arraystretch}{0.95}
\caption{Cohort characteristics for BuddyBench real-data tasks.}
\label{tab:cohort-summary}
\resizebox{\columnwidth}{!}{%
\begin{tabular}{p{0.30\textwidth}p{0.30\textwidth}p{0.30\textwidth}}
\toprule
\textbf{Characteristic} & \textbf{ND-03} & \textbf{ND-02} \\
\midrule
Study design & Observational longitudinal & 12-week RCT \\
Analysis $n$ & 183 & 83 FAS (86 ITT) \\
Sex ratio (M:F) & 2.4:1 & 13:1 \\
Diagnostic composition & SCD 29.5\%, ASD 27.9\%, high-risk 27.3\%, ADHD 13.1\%, ID 1.1\%, other/missing 1.1\% & ASD/SCD concentrated \\
Drill fill rate & 88.1\% & 32.0\% \\
Primary BuddyBench tasks & T1 (KT), T2 (RecSys) & T3 (Clinical), T4 (Causal) \\
\bottomrule
\end{tabular}}
\end{table}

The ND-02 full-analysis set ($n = 83$) was derived from the 86-participant ITT sample after excluding three participants who did not meet FAS inclusion criteria. The non-response label for Task~3 is assigned when a participant shows no improvement or any deterioration across all three primary outcome endpoints (CGI-S/I, SRS-2, VABS), yielding a positive rate of 0.747.

\subsection{Participant-Split Rationale}

All real-data evaluations use a participant-level split: each participant appears in exactly one outer fold and is never shared between training and test portions of the same fold. This design prevents two leakage paths specific to educational and clinical AI benchmarks.

\textbf{Student-level leakage.} In knowledge-tracing and recommendation tasks, a model could memorize per-student patterns if the same student appears in both training and test. Participant-level splits ensure that accuracy reflects generalization to unseen learners, not within-student memorization.

\textbf{Clinical correlation leakage.} In clinical prediction tasks, participants within the same cohort can share unmeasured confounders. Participant-level splits ensure correlated units remain in the same fold rather than leaking across the train-test boundary.

For Tasks~3--4 on ND-02, outer folds are stratified to preserve the 0.747 positive rate across all five folds, preventing artificial AUPRC baseline variation from fold to fold. Fold construction uses seeds 42, 123, 456, 789, and 1024.

\subsection{Re-identification Risk}
\label{app:reidentification}

ND-02 carries elevated re-identification risk due to its small sample size ($n = 83$), extreme sex ratio (13:1 M:F), and concentration in a rare diagnostic population. The following mitigations are applied throughout this paper and in the BuddyBench benchmark release:

\begin{itemize}
\item Per-subgroup performance breakdowns for ND-02 are not reported.
\item No individual-level predictions, fold-membership indicators, or case-level outputs are released.
\item Timestamps are withheld; drill ordering is inferred from column position in the wide data matrix.
\item Real ND-02 and ND-03 data require dual IRB approval (requesting institution + originating study IRB) before a data use agreement can be arranged; BuddyBench-Sim is the only openly available release.
\end{itemize}

Unlike typical attribute inference settings, where a subset of features is observable and a latent sensitive attribute is the inference target, BuddyBench admits no such distinction. Every field in the dataset encodes sensitive pediatric clinical or behavioral information: diagnostic category (ASD or SCD), IQ, age, sex, standardized assessment scores (VABS-II, SRS-2, BuddyPlan), and drill-level response patterns. Re-identification risk is therefore the central privacy concern, and the disclosure audit focuses on whether a synthetic record can be matched to a specific real participant rather than whether a particular attribute can be inferred.

For BuddyBench-Sim, membership-inference accuracy against the real cohort was measured at $\leq 0.52$ (near chance) using a shadow-model classifier trained to distinguish real from synthetic records on held-out samples, and the minimum group size across all reported demographic-diagnostic-severity groupings satisfies $k \geq 5$, consistent with standard $k$-anonymity thresholds for synthetic clinical data.

\section{Statistical Analysis Details}
\label{app:statistics}

Tasks~1--4 use participant-split evaluation to avoid participant leakage on the real-data benchmarks. For Task~1, the appendix reports the current participant-split KT leaderboards for both real and synthetic cohorts. Unless a caption says otherwise, a plain $a \pm b$ entry in the task tables denotes mean $\pm$ fold-wise SD over the reported evaluations for that row; confidence intervals are reported only when explicitly labeled. For Task~4, the main table and appendix leaderboard follow this same rule and compare only residualized-R-loss rows because that metric is shared across the reported estimators and endpoints.

\begin{table}[!htbp]
\centering
\scriptsize
\setlength{\tabcolsep}{2pt}
\renewcommand{\arraystretch}{0.94}
\caption{Uncertainty-reporting ledger for the main task tables.}
\label{tab:uncertainty-ledger}
\resizebox{\columnwidth}{!}{%
\begin{tabular}{lcl}
\toprule
\textbf{Task} & \textbf{Folds} & \textbf{Evaluation note} \\
\midrule
T1 KT & 5 & Subject-fold outer-fold SD \\
T2 RecSys & 5 & Warm-start five-fold SD \\
T3 Clinical & 5 & Nested-CV outer-fold SD \\
T4 Causal & 5 & Residualized R-loss fold-wise SD \\
\bottomrule
\end{tabular}}
\end{table}

\subsection{Confidence Interval Methodology}
\label{app:stats-ci}

Where reported, confidence intervals use bias-corrected accelerated (BCa) bootstrap with $B = 1000$ resamples. In this paper, BCa is used only for results explicitly labeled as 95\% confidence intervals in bracketed form; it is not the interpretation of unlabeled $\pm$ tables. The BCa method corrects for both bias and skewness in the bootstrap distribution:
\begin{align}
\alpha_1 &= \Phi\!\left(\hat{z}_0 + \frac{\hat{z}_0 + z_{\alpha/2}}{1 - \hat{a}(\hat{z}_0 + z_{\alpha/2})}\right) \\
\alpha_2 &= \Phi\!\left(\hat{z}_0 + \frac{\hat{z}_0 + z_{1-\alpha/2}}{1 - \hat{a}(\hat{z}_0 + z_{1-\alpha/2})}\right)
\end{align}
where $\hat{z}_0$ is the bias-correction factor and $\hat{a}$ is the acceleration parameter estimated via jackknife.

\subsection{Multiple Testing Correction}
\label{app:stats-mht}

Benjamini-Hochberg false discovery rate (FDR) control at $q = 0.05$ is applied across all model-metric combinations within each task:
\begin{enumerate}
\item Order p-values: $p_{(1)} \leq p_{(2)} \leq \cdots \leq p_{(m)}$
\item Find the largest $i$ such that $p_{(i)} \leq \frac{i}{m} q$
\item Reject hypotheses $H_{(1)}, \ldots, H_{(i)}$
\end{enumerate}

\paragraph{Task~1 significance results.}
For pyKT-framework models with complete five-fold runs under the F3 subject-split protocol, each model's fold-level AUC vector was compared against the best model (\texttt{PEBG}, mean AUC $= 0.723$) using a paired $t$-test across the five outer folds; resulting $p$-values were corrected with BH-FDR ($q{=}0.05$).
Only \texttt{IEKT} ($p = 0.123$, BH-FDR corrected) was not rejected, forming a two-model competitive cluster with \texttt{PEBG}.
All remaining 14 models were significantly lower than \texttt{PEBG} ($p \leq 0.006$ after BH-FDR correction).

\paragraph{Task~3 significance results.}
For all 14 non-baseline models evaluated on ND-02 under the \texttt{v1\_only}/F2 leakage-safe protocol, each model's five-fold outer AUPRC vector was compared against the best model (\texttt{tabm}, mean AUPRC $= 0.860$) using a paired $t$-test with BH-FDR correction.
Only \texttt{random\_forest} was rejected ($p = 0.0024$, BH-FDR $q{=}0.05$); all other models were statistically indistinguishable from the best, reflecting the low power of five outer folds with $n{=}83$ participants.

\subsection{Effect Size Interpretation}
\label{app:stats-effect}

\begin{table}[!htbp]
\centering
\scriptsize
\setlength{\tabcolsep}{4pt}
\renewcommand{\arraystretch}{0.95}
\caption{Cohen's $d$ effect-size guide for clinical benchmark comparisons.}
\label{tab:effect-sizes}
\begin{tabular*}{\columnwidth}{@{\extracolsep{\fill}}p{0.25\columnwidth}c p{0.42\columnwidth}@{}}
\toprule
\textbf{Magnitude} & \textbf{Cohen's $d$} & \textbf{Clinical interpretation} \\
\midrule
Trivial & 0.00 - 0.19 & Not clinically meaningful \\
Small   & 0.20 - 0.49 & Minimal clinical significance \\
Medium  & 0.50 - 0.79 & Moderate clinical significance \\
Large   & 0.80 - 1.29 & Strong clinical significance \\
Very large & $\geq 1.30$ & Exceptional clinical significance \\
\bottomrule
\end{tabular*}
\end{table}

\section{Hyperparameter Tuning Protocol}
\label{app:tuning}

Task~3 uses nested cross-validation: an outer 5-fold stratified split and an inner 3-fold stratified split for AUPRC tuning. All candidate configurations are evaluated solely on inner-fold validation data; the held-out outer fold sees only the single best inner-fold configuration. This prevents optimistic bias from hyperparameter leakage into test estimates. Table~\ref{tab:tuning-budget} summarizes the search budget. Task~3 covers 20 real ND-02 models with AUPRC as the primary objective, and Table~\ref{tab:t3-search-space} lists representative search spaces.

\begin{table}[!htbp]
\centering
\scriptsize
\setlength{\tabcolsep}{4pt}
\renewcommand{\arraystretch}{0.95}
\caption{Hyperparameter tuning budget for Task~3.}
\label{tab:tuning-budget}
\begin{tabular*}{\columnwidth}{@{\extracolsep{\fill}}lcccc@{}}
\toprule
\textbf{Task} & \textbf{Models} & \shortstack{\textbf{Outer}\\\textbf{folds}} & \shortstack{\textbf{Inner}\\\textbf{folds}} & \textbf{Objective} \\
\midrule
Task~3 & 20 & 5 & 3 & AUPRC \\
\bottomrule
\end{tabular*}
\end{table}

\begin{table}[!htbp]
\centering
\scriptsize
\setlength{\tabcolsep}{3pt}
\renewcommand{\arraystretch}{0.94}
\caption{Representative Task~3 hyperparameter search spaces, part 1.}
\label{tab:t3-search-space}
\resizebox{\columnwidth}{!}{%
\begin{tabular}{lll}
\toprule
\textbf{Family} & \textbf{Hyperparameter} & \textbf{Range / set} \\
\midrule
Linear & $\alpha$ / C & $[10^{-3}, 10^{2}]$ \\
       & l1\_ratio & $[0.0, 1.0]$ \\
       & max\_iter & \{1000, 5000\} \\
\midrule
Tree & learning\_rate & $[0.01, 0.30]$ \\
     & depth / max\_depth & \{4, 6, 8, 10\} \\
     & n\_estimators & \{100, 300, 500\} \\
\bottomrule
\end{tabular}}
\end{table}

\begin{table}[!htbp]
\centering
\scriptsize
\setlength{\tabcolsep}{3pt}
\renewcommand{\arraystretch}{0.94}
\caption{Representative Task~3 hyperparameter search spaces, part 2.}
\label{tab:t3-search-space-continued}
\resizebox{\columnwidth}{!}{%
\begin{tabular}{lll}
\toprule
\textbf{Family} & \textbf{Hyperparameter} & \textbf{Range / set} \\
\midrule
Neural & d\_model & \{32, 64, 128\} \\
       & num\_heads & \{2, 4, 8\} \\
       & dropout & $[0.1, 0.5]$ \\
       & learning\_rate & $[10^{-4}, 10^{-3}]$ \\
\midrule
Foundation & n\_estimators & \{8, 16, 32\} \\
           & softmax\_temperature & $[0.5, 2.0]$ \\
\bottomrule
\end{tabular}}
\end{table}

\section{Task~2/3/4 Supplementary Notes}
\label{app:t245-raw-supplement}

This section summarizes the Task~2-4 appendix tables. Real-data and BuddyBench-Sim results are reported separately, and non-finite rows are left unreported rather than imputed.

\subsection{Task~2 Recommendation}
\label{app:t245-task2}

Table~\ref{tab:t2-recsys-extended} is the primary real ND-03 leaderboard for the warm-start F3 recommendation benchmark and should be used when comparing methods on the paper's primary offline ranking task. Tables~\ref{tab:synthetic-t2} and~\ref{tab:synthetic-t2-continued} serve a different purpose: they report retention and transfer behavior on BuddyBench-Sim, making them a public-interface diagnostic rather than a substitute real-data leaderboard.

\subsection{Task~3 Clinical Prediction}
\label{app:t245-task3}

Task~3 real ND-02 clinical prediction uses nested 5-fold outer CV with 3-fold inner tuning, F2 pre-treatment-only features, and AUPRC as the tuning objective. Within that setup, Table~\ref{tab:t3-ablation} provides the feature-tier evidence for the primary F2 reporting point, and Table~\ref{tab:t3-response-nd02} is the primary real-data leaderboard. Tables~\ref{tab:t3-response-synthetic-full} and~\ref{tab:t3-response-synthetic-verify} then provide auxiliary BuddyBench-Sim F3 evidence for executability and ranking retention at full-size and rough fixed-size parity, respectively. Both refreshed synthetic blocks are sourced from the BuddyBench-Sim clinical-prediction data, and rows with unavailable feature dependencies are excluded rather than imputed.

\subsection{Task~4 Causal Inference}
\label{app:t245-task4}

Task~4 real ND-02 rows use residualized R-loss, and Table~\ref{tab:t4-causal-extended} is the full real-data leaderboard for comparing causal method families under the paper's small-RCT protocol. PEHE is omitted there because individual counterfactuals are unobservable in the real trial. Table~\ref{tab:synthetic-t4} is therefore a supplementary synthetic-only comparison that adds PEHE where the simulation makes it identifiable, rather than a direct replacement for the real-data benchmark. The synthetic F3 rows come from the BuddyBench-Sim causal-simulation data and cover the tier-5 estimators evaluated under the released benchmark environment.

\section{Sample-Size Scaling Analysis (T3 \& T4)}
\label{app:scaling}

\begin{figure}[h]
\centering
\includegraphics[width=\linewidth]{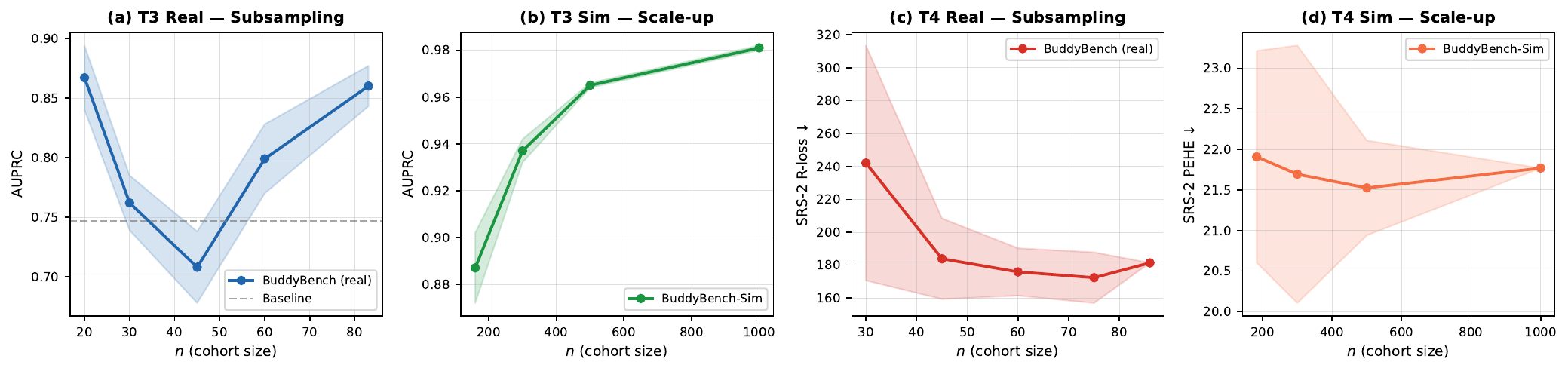}
\caption{Sample-size scaling analysis for T3 (panels a-b) and T4 (panels c-d). \textbf{(a)} T3 real-data subsampling: best-model AUPRC vs.\ cohort size $n \in \{20,30,45,60,83\}$; dashed line = random baseline (0.747). \textbf{(b)} T3 BuddyBench-Sim scale-up: AUPRC vs.\ $n \in \{160,300,500,1000\}$, showing monotone improvement from $0.887$ to $0.981$. \textbf{(c)} T4 real-data subsampling: best-model SRS-2 residualized R-loss (lower is better) vs.\ $n \in \{30,45,60,75,86\}$. \textbf{(d)} T4 BuddyBench-Sim scale-up: SRS-2 PEHE (lower is better, uniquely computable on synthetic data) vs.\ $n \in \{183,300,500,1000\}$. Each point is the best-model value averaged over three random stratified seeds; shaded bands show $\pm$1 SD.}
\label{fig:scaling-combined}
\end{figure}

\paragraph{Real-data subsampling (panels a and c).}
Figure~\ref{fig:scaling-combined} panels (a) and (c) report AUPRC (T3) and SRS-2 residualized R-loss (T4) as a function of real ND-02 cohort size, obtained by stratified subsampling at $n \in \{20, 30, 45, 60, 83\}$ (T3) and $n \in \{30, 45, 60, 75, 86\}$ (T4) with three random seeds each. For T3, AUPRC at the full $n = 83$ reaches $0.860 \pm 0.017$, while subsampling to $n = 45$ reduces it to $0.708 \pm 0.030$, a 15-point drop that confirms the benchmark operates well below its model-family ceiling. The apparent non-monotone pattern at $n = 20$ (AUPRC $0.867$) is consistent with high label-imbalance variance: the positive rate of 0.747 leaves fewer than six negative examples at $n = 20$, making stratified subsampling unstable. For T4, mean best R-loss on the SRS-2 endpoint falls from $242 \pm 71$ at $n = 30$ to $172 \pm 15$ at $n = 75$, a 29\% reduction, demonstrating that the estimators are data-starved rather than misspecified. Both curves show no sign of saturation at the available cohort sizes, directly defending the observed low discriminative power in T3 and the high absolute R-loss in T4 as consequences of the inherently small pediatric RCT regime rather than of benchmark design or model selection.

\paragraph{BuddyBench-Sim scale-up (panels b and d).}
Panels (b) and (d) show the unique advantage of BuddyBench-Sim: because synthetic cohorts can be generated at arbitrary $n$, scale-up experiments are possible well beyond the real-data ceiling. For T3, Sim AUPRC improves monotonically from $0.887 \pm 0.015$ at $n = 160$ to $0.981 \pm 0.001$ at $n = 1000$, with seed variance collapsing as sample size grows. Already at $n = 160$, roughly twice the real cohort, the synthetic benchmark exceeds real full-cohort performance, confirming that the data-quantity bottleneck rather than model capacity limits T3 real performance. For T4, Sim PEHE on the SRS-2 endpoint, a metric uniquely computable for synthetic data where true counterfactuals are known from the DGP, ranges from $21.9 \pm 1.3$ at $n = 183$ to $21.5 \pm 0.6$ at $n = 500$. The modest improvement range reflects near-optimal PEHE even at the smallest Sim size, consistent with limited treatment-effect heterogeneity in the SRS-2 simulation. Together, these scale-up curves establish BuddyBench-Sim as a scalable development interface: researchers can pre-screen model families at large $n$ on synthetic data before submitting to the restricted real-data leaderboard.

\section{Computational Infrastructure}
\label{app:computation}

\textbf{Hardware:}
\begin{itemize}
\item All experiments were run on a single workstation with one NVIDIA RTX 5090 GPU.
\end{itemize}

\textbf{Software Environment:}
\begin{itemize}
\item Operating System: Ubuntu 22.04 LTS
\item Python: 3.11
\item Key libraries: \texttt{PyTorch} 2.1, \texttt{scikit-learn} 1.4, \texttt{CatBoost} 1.2, \texttt{EconML} 0.15
\end{itemize}

\textbf{Approximate Runtimes:}
\begin{itemize}
\item Task~1 KT models: 2-8 hours training per model (original sweep)
\item Task~2 RecSys benchmark: 1-4 hours per model
\item Task~3 clinical-prediction benchmark: $<$1 hour for full suite
\item Task~4 causal benchmark: 1-6 hours per tier
\end{itemize}

\section{Dataset Documentation}
\label{app:datasheet}

This section provides a structured record for BuddyBench-Sim following the Datasheets for Datasets framework \citep{gebru2021datasheets}.

\paragraph{Motivation.}
BuddyBench-Sim was created to support reproducible evaluation of computational methods for ASD/SCD digital therapeutics without exposing real patient data. The goal is a standardized, openly available development interface that preserves the statistical structure of the real cohorts while eliminating re-identification risk.

\paragraph{Composition.}
BuddyBench-Sim contains 1,000 synthetic records generated by a PI-VAE model trained on the combined ND-02 and ND-03 real cohorts. Task-specific files are provided for T1 (knowledge tracing), T2 (next-drill recommendation), T3 (clinical prediction), and T4 (causal simulation). The dataset contains no real patient records and no personally identifiable information. Labels for T3 and T4 are derived from the synthetic data-generating process; no human annotation was required.

\paragraph{Collection Process.}
Records were generated via PI-VAE, which models the joint distribution of real cohort features under privacy-preserving regularization. Pairwise Spearman feature-correlation alignment between synthetic and real distributions is $\geq 0.968$; record-level clinical validity (proportion satisfying clinical plausibility constraints) is 0.990. No real records are included in the release.

\paragraph{Preprocessing.}
Synthetic records were post-processed to enforce the same feature-pipeline constraints as the real benchmark: F0-F3 feature tier construction, leakage constraints (no post-treatment variables for T3-T4), and the Task~1 tokenization protocol. Rows with dependency-unavailable fields under the maintained benchmark environment are excluded from task evaluation files rather than imputed.

\paragraph{Uses.}
BuddyBench-Sim is intended for (1) open development and continuous-integration testing of benchmark code, (2) preliminary model screening before requesting real-data access, and (3) benchmark methodology research. It is \emph{not} intended for clinical validation, epidemiological inference, or subgroup-level claims about ASD/SCD populations. Users who draw clinical conclusions from BuddyBench-Sim results without real-data validation are misusing the resource.

\paragraph{Distribution.}
BuddyBench-Sim is released under the Creative Commons Attribution-NonCommercial 4.0 International (CC~BY-NC~4.0) license and hosted in the benchmark repository. Real ND-02 and ND-03 data are not included and are not publicly available. Access requires IRB approval from both the requesting researcher's institution and the originating study's IRB, after which a data use agreement (DUA) can be arranged with the originating institution.

\paragraph{Maintenance.}
The dataset is maintained alongside the benchmark codebase under a versioned release scheme. Issues should be filed in the public repository. The authors commit to maintaining compatibility with the released evaluation pipeline for at least two years following the initial public release.

\section{Reproducibility Checklist}
\label{app:reproducibility}

\textbf{Data:}
\begin{itemize}
\item BuddyBench-Sim publicly available under CC~BY-NC~4.0 license
\item Real data access requires dual IRB approval (requesting institution + originating study IRB); access protocol available upon request to the corresponding author
\item Data preprocessing code provided in the benchmark repository
\item Train/validation/test splits standardized and released with BuddyBench-Sim
\end{itemize}

\textbf{Code:}
\begin{itemize}
\item Complete benchmark implementation publicly available
\item Baseline model implementations for all registered models across T1-T4
\item Evaluation metrics and statistical analysis code (including BCa bootstrap)
\item Synthetic data generation pipeline
\end{itemize}

\textbf{Experimental Details:}
\begin{itemize}
\item Hyperparameter settings documented in the released benchmark materials for each task
\item Random seeds: 42, 123, 456, 789, 1024 (where the runner supports multi-seed evaluation)
\item Cross-validation procedure: participant-split for Tasks~1--4 in the real-data tables; BuddyBench-Sim columns are reported where available
\item Extended benchmark tables are traceable to the archived benchmark outputs
\end{itemize}

\end{document}